\documentclass[lettersize,journal]{IEEEtran}
\usepackage{amsmath,amsfonts}
\usepackage{algorithmic}
\usepackage{algorithm}
\usepackage{array}
\usepackage[caption=false,font=normalsize,labelfont=sf,textfont=sf]{subfig}
\usepackage{textcomp}
\usepackage{stfloats}
\usepackage{url}
\usepackage{verbatim}
\usepackage{graphicx}
\usepackage{cite}
\usepackage{multirow}
\usepackage{booktabs}
\usepackage{graphicx}
\usepackage{color}
\usepackage{overpic}
\usepackage{rotating}
\usepackage{pifont}
\begin{document}
\title{A Feature Reuse Framework with Texture-adaptive Aggregation for Reference-based Super-Resolution}

\author{Xiaoyong Mei,~%\IEEEmembership{Member,~IEEE,}
        Yi Yang,~%\IEEEmembership{Fellow,~OSA,}
        Ming~Li*,~\IEEEmembership{Member,~IEEE,}
        Changqin Huang,~\IEEEmembership{Member,~IEEE,}\\
        Kai Zhang,~\IEEEmembership{Member,~IEEE,}
        Pietro Li\'o
\thanks{X. Mei, Y. Yang, M. Li, and C. Huang are with the Key Laboratory of Intelligent Education Technology
and Application of Zhejiang Province, Zhejiang Normal University, Jinhua, China (e-mail: cdmxy@zjnu.edu.cn; yangyi@zjnu.edu.cn; mingli@zjnu.edu.cn; cqhuang@zju.edu.cn).}% <-this % stops a space
\thanks{K. Zhang is with the Computer Vision Lab, ETH Zurich, Switzerland (e-mail: cskaizhang@gmail.com).}
\thanks{P. Li\'o  is with the Department of Computer Science and Technology, University of Cambridge, UK (e-mail: Pietro.Lio@cl.cam.ac.uk).}

\thanks{*Corresponding author}
\markboth{Submitted to IEEE TIP
}%
{Shell \MakeLowercase{\textit{et al.}}:}}


\maketitle

\begin{abstract}
Reference-based super-resolution (RefSR) has gained considerable success in the field of super-resolution with the addition of high-resolution reference images to reconstruct low-resolution (LR) inputs with more high-frequency details, thereby overcoming some limitations of single image super-resolution (SISR). Previous research in the field of RefSR has mostly focused on two crucial aspects. The first is accurate correspondence matching between the LR and the reference (Ref) image. The second is the effective transfer and aggregation of similar texture information from the Ref images. Nonetheless, an important detail of perceptual loss and adversarial loss has been underestimated, which has a certain adverse effect on texture transfer and reconstruction. In this study, we propose a \emph{feature reuse} framework that guides the step-by-step texture reconstruction process through different stages, reducing the negative impacts of perceptual and adversarial loss. The \emph{feature reuse} framework can be used for any RefSR model, and several RefSR approaches have improved their performance after being retrained using our framework. Additionally, we introduce a single image feature embedding module and a texture-adaptive aggregation module. The single image feature embedding module assists in reconstructing the features of the LR inputs itself and effectively lowers the possibility of including irrelevant textures. The texture-adaptive aggregation module dynamically perceives and aggregates texture information between the LR inputs and the Ref images using dynamic filters. This enhances the utilization of the reference texture while reducing reference misuse. The source code is available at https://github.com/Yi-Yang355/FRFSR.
\end{abstract}


\section{Introduction}
\IEEEPARstart{S}{ingle} Image Super-Resolution (SISR) involves generating a high-resolution image with high-frequency information from a low-resolution (LR) input. The practical significance of SISR in various contexts such as medical imaging and surveillance is notable. Based on the optimization criteria, the approaches of SISR can be divided into two categories. One approach optimizes pixel-level errors such as mean squared error (MSE) and mean absolute error (MAE), potentially resulting in images that are too smooth, and the other approach involves visual perception-based errors such as perceptual loss and adversarial loss. The latter results in images with better visual effects and greater alignment to human visual perception but may produce artifacts and unrealistic textures. These approaches face the inherent problem of SISR - the ill-posed nature of the problem - because different high-resolution images can be degraded to the same low-resolution image~\cite{closed-loop,dip}. Reference-based super-resolution (RefSR) alleviates the inherent problem of SISR to a certain extent by using an additional high-resolution reference (Ref) image to transfer relevant textures and achieve super-resolution. Methods of obtaining relevant Ref images are varied and include web search and video frames. RefSR has two primary limitations that compromise its performance. The first one is accurately finding the correspondence between the LR and Ref. Some existing methods address this through spatial alignment, such as CrossNet~\cite{crossnet}, which utilizes optical flow estimation to align LR and Ref, and SSEN~\cite{ssen}, which employs deformable convolutions to learn adaptive LR and Ref alignment. Other methods, such as SRNTT~\cite{srntt}, TTSR~\cite{ttsr} adopt dense patch matching algorithms for patch matching to find corresponding matches, whereas MASA~\cite{masa} employs a coarse-to-fine matching approach for reducing computational requirements. However, obtaining accurate matching is challenging due to differences in resolution and texture distribution. $C^2$-Matching~\cite{c2} uses knowledge distillation and contrastive learning to train a feature extractor, and a combination of patch matching and deformable convolution to improve the accuracy of correspondence matching. However, deformable convolution~\cite{dcn, dcnv2} still encounters difficulties in aligning features at long distances. The second challenge is effectively transferring texture features. TTSR proposes a cross-scale feature integration module that conveys texture information using multiple texture transformers in a stacked manner, whereas MASA uses a spatial adaptive module to remap the aligned Ref feature distribution, ensuring robustness to different color and brightness distributions. Additionally, DATSR~\cite{datsr} replaces the traditional ResBlock with the Swin-Transformer~\cite{swint}, resulting in considerable improvements in model performance.

Although deformable convolution is capable of learning implicit alignment between feature maps LR and Ref, it still faces challenges in aligning distant features. Furthermore, existing RefSR methods effectively prioritize aggregating textures over reconstructing their own textures. It is also important to note that during the feature aggregation process, the ResBlock treats all pixel features equally, resulting in the introduction of irrelevant textures from the Ref image. Even with DATSR replacing ResBlock with Swin-Transformer, the window self-attention calculation will noticeably increase the parameters and runtime.

To address these three issues, we first do not make any modifications to the deformable convolution, but instead shuffle the reference image, thereby indirectly increasing the distance between similar features, increasing the training difficulty and improving performance; secondly, inspired by TADE~\cite{tade}, we use single-image feature embedding to assist the LR inputs to self-reconstruct their features while mitigating the introduction of irrelevant textures. Finally, we introduce a new feature aggregation module, namely Dynamic ResBlock (DRB). Specifically, the DRB module adds a group of decoupled filters to the residual block, which can aware texture information in both the spatial and channel domains, and then adaptively aggregate relevant textures, further reducing the introduction of irrelevant information such as noise, wrong textures, etc., using an efficient enhanced spatial attention mechanism (ESA) to enhance relevant texture information.

In addition to the aforementioned points, most previous works overlook a crucial fact: the increase in perceptual loss and adversarial loss adversely affects the texture transfer and reconstruction effects. To fully utilize the texture transfer and reconstruction abilities of the reconstruction loss-trained model, we propose a feature reuse framework. In the process of ${{\cal L}^{\textit{rec}}}$+${{\cal L}^{\textit{per}}}$+${{\cal L}^{\textit{adv}}}$ training and testing, we feed back the features trained only with ${{\cal L}^{\textit{rec}}}$ to the feature aggregation module. This maneuver effectively diminishes the impact of perceptual and adversarial losses on texture transfer and reconstruction. In summary, this paper's primary contributions are:
\begin{itemize}
\item{We introduce a feature reuse framework that significantly reduces texture reconstruction degradation resulting from perceptual loss and adversarial loss. We apply this framework to various RefSR methods, which have shown consistent improvements in performance.}
\item{To enhance the reconstruction of LR's self-texture and maintain texture relevance, we utilize a single-image feature reconstruction module. Unlike the approach used by~\cite{tade}, we exclude feature upsampling and final image reconstruction processes in this module, and focus solely on embedding the LR's own reconstructed features into the aggregation process.}
\item{We designed a dynamic residual block and introduced it into the texture adaptive module. This block applies dynamic filters and enhanced spatial attention to selectively perceive and transfer textures from the Ref image. This approach adaptively reduces the likelihood of introducing incorrect textures.}
\item{Our method achieved state-of-the-art (SOTA) performance in multiple benchmarks, demonstrating significant improvements in robustness to unrelated reference images and long-range feature alignment. Notably, even without the single-image feature reconstruction module, our method still achieved SOTA performance in CUFED5.}
\end{itemize}
\section{Related Works}
\subsection{Single Image Super-Resolution}
Single image super-resolution (SISR) aims to input a single LR image and reconstruct it to an image with high-frequency details. Before the emergence of deep learning, traditional methods such as various interpolation methods were usually used. With SRCNN~\cite{srcnn} first using deep learning methods to perform super-resolution, deep learning-based super-resolution began to appear in large numbers. Later, ResNet~\cite{resnet} appeared, which deepened the network layers. EDSR~\cite{edsr}, CARN~\cite{carn} and other methods added residual structure in super-resolution models, thus improving the performance of super-resolution. After this, the attention mechanism merged, which can make the network selectively focus on some features and appropriately ignore unnecessary features. RCAN~\cite{rcan} was the first to apply the attention mechanism to super-resolution. Additionally, the game theory approach used by GAN~\cite{gan} has enabled GAN-based super-resolution models, such as SRGAN~\cite{srgan}, ESRGAN~\cite{esrgan}, RankSRGAN~\cite{rankgan}, AMPRN~\cite{AMPRN}, and Real-ESRGAN~\cite{real-esrgan} to deliver enhanced perceptual quality in produced images. Recently, SRGAT~\cite{srgat} used the graph attention network to help LR recover additional textures from neighboring patches. TDPN\cite{tdpn} utilizes a texture and detail-preserving network that preserves texture and detail while the features are reconstructed. However, the SISR problem is ill-posed, with low-resolution (LR) and super-resolution (SR) having a one-to-many relationship.
\subsection{Reference-based Image Super-Resolution}
The biggest difference between RefSR and SISR is that the former has an additional high-resolution Ref image. The RefSR can transfer texture details from the Ref image to LR to help LR reconstruction, and these texture details should be similar to the ground truth (GT). CrossNet~\cite{crossnet} twists the reference image and LR to align them through the flow estimation network. SSEN~\cite{ssen} uses deformable convolution~\cite{dcn,dcnv2} to align LR and Ref images. Both of these methods are implicit alignment, and some work performs feature matching between LR and Ref to achieve explicit alignment. SRNTT~\cite{srntt} enumerates patches to transfer multi-scale reference features. TTSR~\cite{ttsr} introduces the Transformer architecture to more reasonably transfer reference features by combining soft and hard attention. MASA~\cite{masa} uses a matching method from coarse to fine to reduce the computational complexity and a spatial adaptive module is used to make the transferred texture closer to GT. However, due to the resolution gap between the LR and Ref image, the matching performance is affected. $C^2$-Matching~\cite{c2} introduces knowledge distillation and contrastive learning methods, which greatly improve the matching robustness between LR and Ref. WTRN~\cite{wtrn} utilizes the benefits of wavelet transformation to categorize features into high-frequency and low-frequency sub-bands, which facilitates the transfer of texture patterns with more effectiveness. TADE~\cite{tade} uses a decoupling framework, which divides RefSR into two parts: super-resolution and texture migration, which alleviates the two problems of reference-underuse and reference-misuse. However, it does not take into consideration the lack of detailed textures in the super-resolution image, which results in inaccurate matching between SR and Ref. DATSR~\cite{datsr} uses the Swin-Transformer~\cite{swint} to replace the traditional ResBlock for feature aggregation. Recently, RRSR~\cite{rrsr} implemented a reciprocal learning strategy, thereby strengthening the learning of the model. Reviewing the existing research findings, it can be seen that first, the existing methods do not fully take into consideration the textural dissimilarities between LR and Ref, so it is still inevitable that irrelevant textures are introduced in the texture transfer process. Second, existing studies have focused on improving the accuracy of matching and the ability of texture transfer, but few studies have focused on the texture detail reconstruction of LR itself. Third, no one has noticed that adding perceptual loss and adversarial loss will lead to a decline in the texture reconstruction effect. To address the aforementioned issues, we propose a dynamic residual block (DRB) to perceive texture information, adaptively transfer and aggregate relevant textures and suppress irrelevant textures and reconstruct their own features by embedding single-image feature reconstruction LR features. In addition, we propose a feature reuse framework to improve the texture reconstruction effect under perceptual loss and adversarial loss supervision.
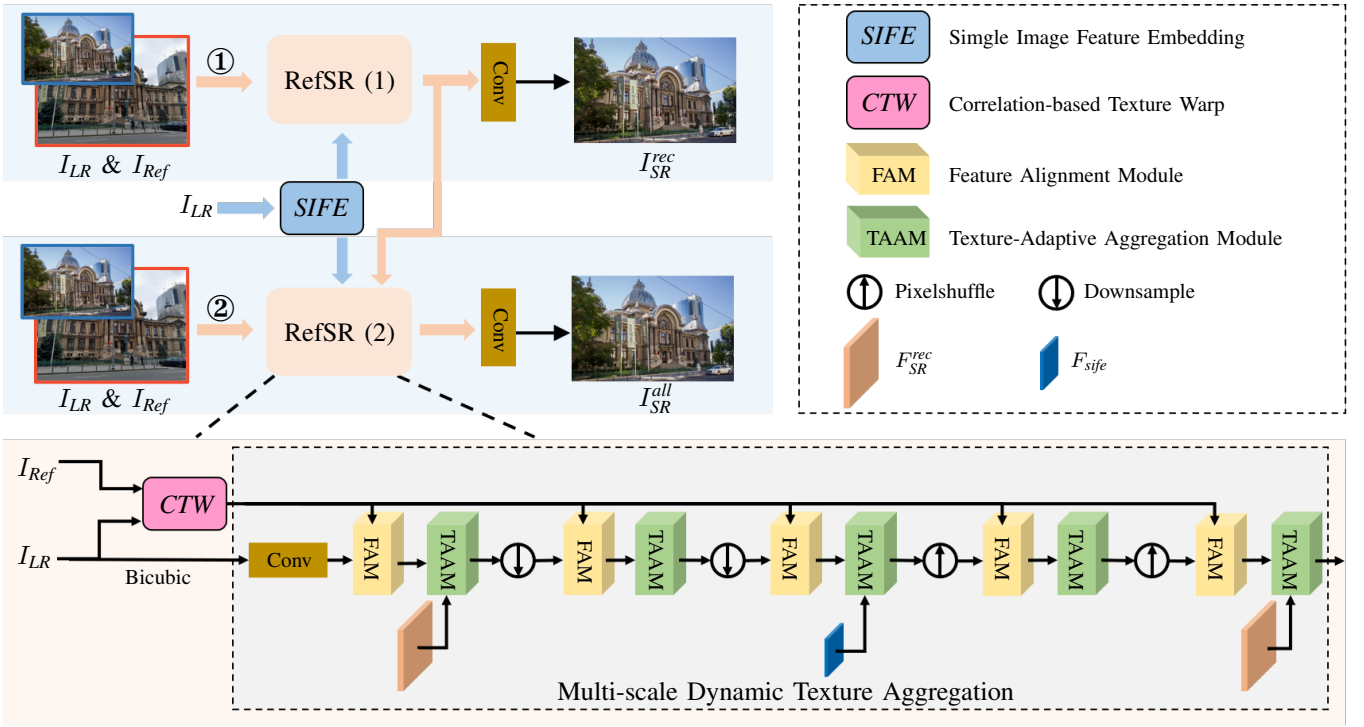
\begin{figure*}[!t]
\begin{center}
\centering
\begin{overpic}[width=0.99\textwidth]{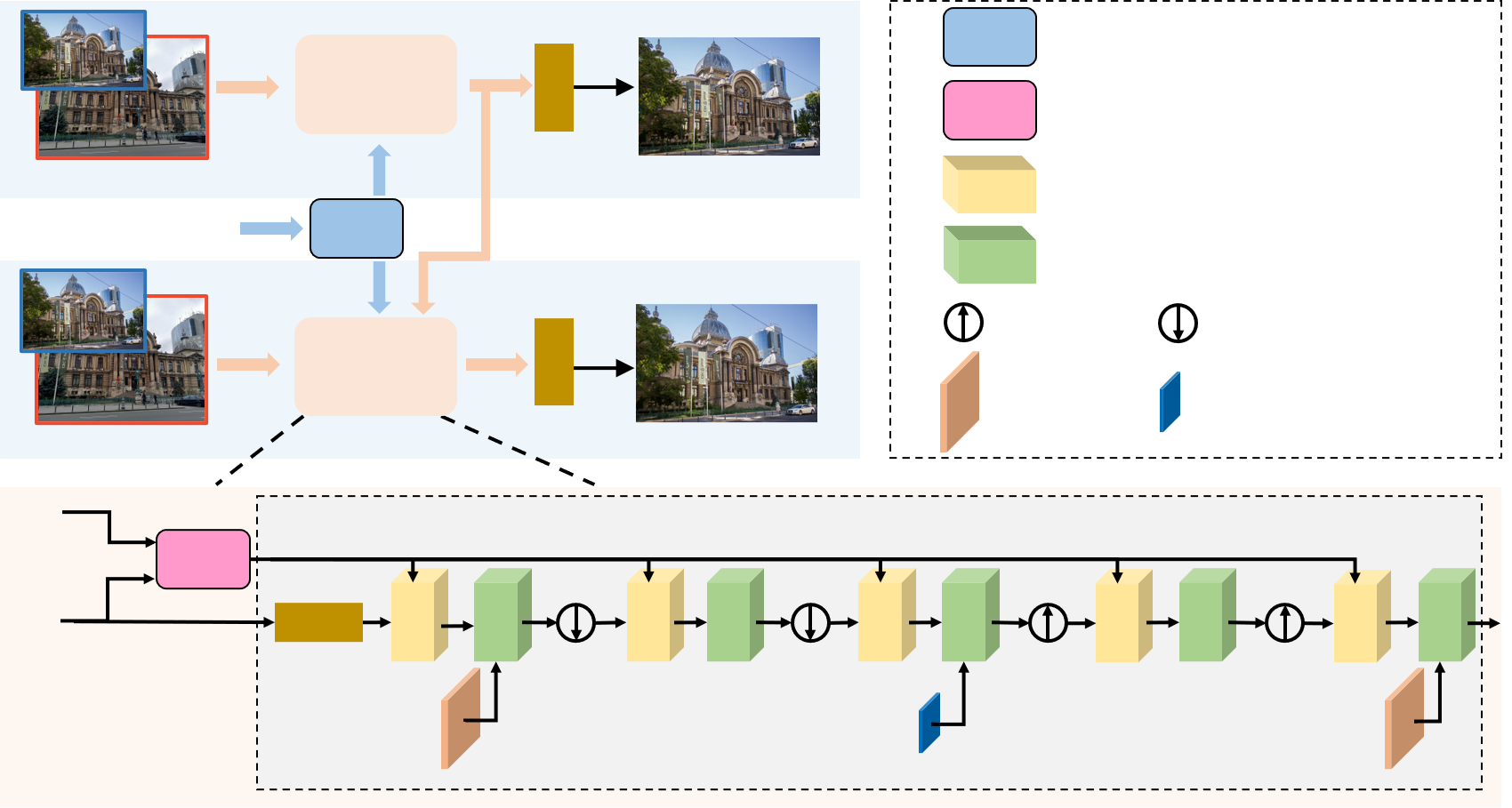}
\put(4,41){\color{black}{\normalsize $I_{\textit{LR}}$ \& $I_{\textit{Ref}}$}}
\put(13,38){\color{black}{\normalsize $I_{\textit{LR}}$}}
\put(4,23.6){\color{black}{\normalsize $I_{\textit{LR}}$ \& $I_{\textit{Ref}}$}}
\put(15,30){\color{black}{\Large \ding{173}}}
\put(15,48.3){\color{black}{\Large \ding{172}}}
\put(20.8,47.2){\color{black}{\normalsize RefSR (1)}}
\put(20.8,28.5){\color{black}{\normalsize RefSR (2)}}
\put(21.5,37.8){\color{black}{\normalsize \textit{SIFE}}}
\put(47,41.1){\color{black}{\normalsize $I_{\textit{SR}}^{\textit{rec}}$}}
\put(47,23.7){\color{black}{\normalsize $I_{\textit{SR}}^{\textit{all}}$}}
\put(63.5,50.5){\color{black}{\normalsize \textit{SIFE}}}
\put(70,50.5){\color{black}{\footnotesize Simgle Image Feature Embedding}}
\put(63.5,45.6){\color{black}{\normalsize \textit{CTW}}}
\put(70,45.6){\color{black}{\footnotesize Correlation-based Texture Warp}}
\put(64.3,40.3){\color{black}{\footnotesize FAM}}
\put(70,40.3){\color{black}{\footnotesize Feature Alignment Module}}
\put(63.9,35.6){\color{black}{\footnotesize TAAM}}
\put(70,35.6){\color{black}{\footnotesize Texture-Adaptive Aggregation Module}}
\put(66,31.7){\color{black}{\footnotesize Pixelshuffle}}
\put(80,31.7){\color{black}{\footnotesize Downsample}}
\put(66,26.6){\color{black}{\footnotesize $F_{\textit{SR}}^{\textit{rec}}$}}
\put(79,26.6){\color{black}{\footnotesize $F_{\textit{sife}}$}}
\put(1,18.5){\color{black}{\normalsize $I_{\textit{Ref}}$}}
\put(1,12){\color{black}{\normalsize $I_{\textit{LR}}$}}
\put(9,10.5){\color{black}{\footnotesize Bicubic}}
\put(11.5,15.8){\color{black}{\normalsize \textit{CTW}}}
\put(19.5,11.8){\color{black}{\footnotesize Conv}}
\put(41, 1.9){\color{black}{\normalsize Multi-scale Dynamic Texture Aggregation}}
\begin{turn}{270}
\put(-31,36.3){\color{black}{\footnotesize Conv}}
\put(-49.3,36.3){\color{black}{\footnotesize Conv}}
\put(-14,26.7){\color{black}{\footnotesize FAM}}
\put(-14.5,32.4){\color{black}{\footnotesize TAAM}}
\put(-13.8,42.5){\color{black}{\footnotesize FAM}}
\put(-14.5,47.8){\color{black}{\footnotesize TAAM}}
\put(-14,57.7){\color{black}{\footnotesize FAM}}
\put(-14.5,63.3){\color{black}{\footnotesize TAAM}}
\put(-14,73.4){\color{black}{\footnotesize FAM}}
\put(-14.5,78.9){\color{black}{\footnotesize TAAM}}
\put(-14,89.2){\color{black}{\footnotesize FAM}}
\put(-14.5,94.7){\color{black}{\footnotesize TAAM}}
\end{turn}
\end{overpic}
\end{center}
\caption{The architecture of our FRFSR. We first utilized SIFE to reconstruct the features of the $I_{LR}$, obtaining $F_{sife}$, which was then embedded into two RefSR models. We eliminated the upsampling and image reconstruction process in SIFE. Next, RefSR (1) was trained solely using the reconstruction loss (-$rec$) and then all loss was utilized in training RefSR (2). We fed back $F_{SR}^{rec}$, which RefSR (1) reconstructed, during the process into the feature aggregation process to guide RefSR (2) in retaining more texture features.}
\label{framework}
\end{figure*}
\subsection{Dynamic Weights}
Unlike the weight sharing in conventional convolutions, dynamic filters~\cite{condconv,weightnet,pla_dy_conv,dedyfilter,senet} have content-aware characteristics and are capable of dynamically adjusting and predicting filter weights based on input features. The dynamic weights approach has been successfully applied in various works, such as super-resolution~\cite{deepsum, metasr, ddet}, image deblurring~\cite{dy_deblur}, image denoising~\cite{dy_denoise}, and style transfer~\cite{dy_trans}, because of its powerful representation and content-awareness capabilities. The work in~\cite{rrsr}, which introduces a set of reference-aware filters for selecting reference features to identify the most suitable texture, is strongly related to our study. However, the generation of these filters is computationally expensive due to their deep separable and spatially changing nature, leading to high time consumption. Inspired by~\cite{dedyfilter}, we propose to decouple the spatial and channel domains and use spatial and channel attention to dynamically filter each pixel, extending this to texture-adaptive aggregation.
\section{Proposed Method}
\subsection{Feature Reuse Reconstruction Framework}
Feature reuse~\cite{resnet,densenet,unet,wideresnet,highwaynet} prevents the vanishing gradient issue in deep networks to enhance network learning and parameter efficiency by inputting previous layers' features into subsequent layers. Various computer vision tasks, such as super-resolution~\cite{densesr}, image compression~\cite{compre_reuse}, and image restoration~\cite{gutresor}, utilize the characteristic of feature reuse to enhance the efficiency and effectiveness of their models. Prior studies have shown that SR images which are produced using only reconstruction loss are much more detailed in texture compared to those generated by models that use perceptual and adversarial losses. To overcome this problem, we propose to utilize a pre-trained model that generates SR feature maps with fine textures through reconstruction loss only, and integrate them into the second model trained with all losses to supplement texture reconstruction and accelerate convergence of the second model, as shown in Fig.1. Therefore, we extend feature reuse to the training process of the two models. In summary, we first input the LR image $I_{\textit{LR}}$ and Ref image $I_{\textit{Ref}}$ into the network to obtain $F_\textit{SR}^{\textit{rec}}$, which is then convolved to generate the output image $I_{\textit{SR}}^\textit{rec}$. In this process, we only train the RefSR model with reconstruction loss, consistent with previous RefSR methods.
\begin{gather}
\label{deqn_ex1a}
{F_{\textit{SR}}^{\textit{rec}}} = \textit{Net}_1(I_{\textit{LR}},I_{\textit{Ref}}),\\
{I_{\textit{SR}}^{\textit{rec}}} = \textit{Conv}(F_{\textit{SR}}^{\textit{rec}}).
\end{gather}
At this stage, we have acquired an RefSR network that exhibits impressive texture transfer and reconstruction abilities. However, to produce high-quality perceptual images, supervision using perceptual and adversarial losses is typically required. To further enhance the second network's texture transfer and reconstruction capabilities, we generate $F_\textit{SR}^{\textit{rec}}$ with refined texture details using the first network, and then incorporate this feature map back into the training of the second network. Note that in this process, the first model is only responsible for inference and does not participate in weight updating. The aforementioned process can be represented as follows:
\begin{gather}
\label{deqn_ex1a}
{F_{\textit{SR}}^{\textit{all}}} = \textit{Net}_2(I_{\textit{LR}},I_{\textit{Ref}},F_{\textit{SR}}^{\textit{rec}}), \\
{I_{\textit{SR}}^{\textit{all}}} = \textit{Conv}(F_{\textit{SR}}^{\textit{all}}).
\end{gather}
By utilizing this framework, we are able to obtain two models with identical texture transfer and reconstruction performance. It is significant to note that this framework has the ability to improve the performance of other RefSR methods. In the ablation study, we apply this framework to MASA~\cite{masa} and $C^2$-Matching~\cite{c2}, demonstrating a significant improvement in their performance.
\subsection{Correlation-based Texture Warp}
For the RefSR task, a large part of the work is focused on accurately finding the matching correspondence between the LR image and the Ref image. This is crucial for subsequent texture transfer. The CTW structure is shown in Fig.2. Firstly, we utilized zero-padding for LR images to ensure they remain the same size as the Ref images. Then, we use a parameter-sharing texture encoder to extract the texture features of LR and Ref images and generate $F_{\textit{LR}}^{\textit{tex}}\in{\mathbb{R}^{C{\times}H_{\textit{LR}}{\times}W_{\textit{LR}}}}$, $F_{\textit{Ref}}^{\textit{tex}}\in{\mathbb{R}^{C{\times}H_{\textit{Ref}}{\times}W_{\textit{Ref}}}}$. We keep the texture encoder consistent with~\cite{c2} because its training method of knowledge distillation and contrastive learning alleviates the problem of inaccurate matching between LR and the reference image due to different resolutions, and enhances the robustness of matching. Then, the texture features $F_{\textit{LR}}^{\textit{tex}}$ and $F_{\textit{Ref}}^{\textit{tex}}$ are respectively unfolded into $l1\left( {{H_{\textit{LR}}} \times {W_{\textit{LR}}}} \right)$, $l2\left( {{H_{Ref}} \times {W_{\textit{Ref}}}} \right)$ patches to obtain $\left\{ {{Q_1},{Q_2},{Q_3}, \ldots ,{Q_{l1}}} \right\}$, $\left\{ {{K_1},{K_2},{K_3}, \ldots ,{K_{l2}}} \right\}$. The cosine similarity between $Q_i$ and each patch $K_j$ is calculated using the inner product formula to form the similarity matrix ${{\cal M}_{i,j}}\in \mathbb{R}^{l1}$.
\begin{figure}[!t]
\begin{center}
\centering
\begin{overpic}[width=0.49\textwidth]{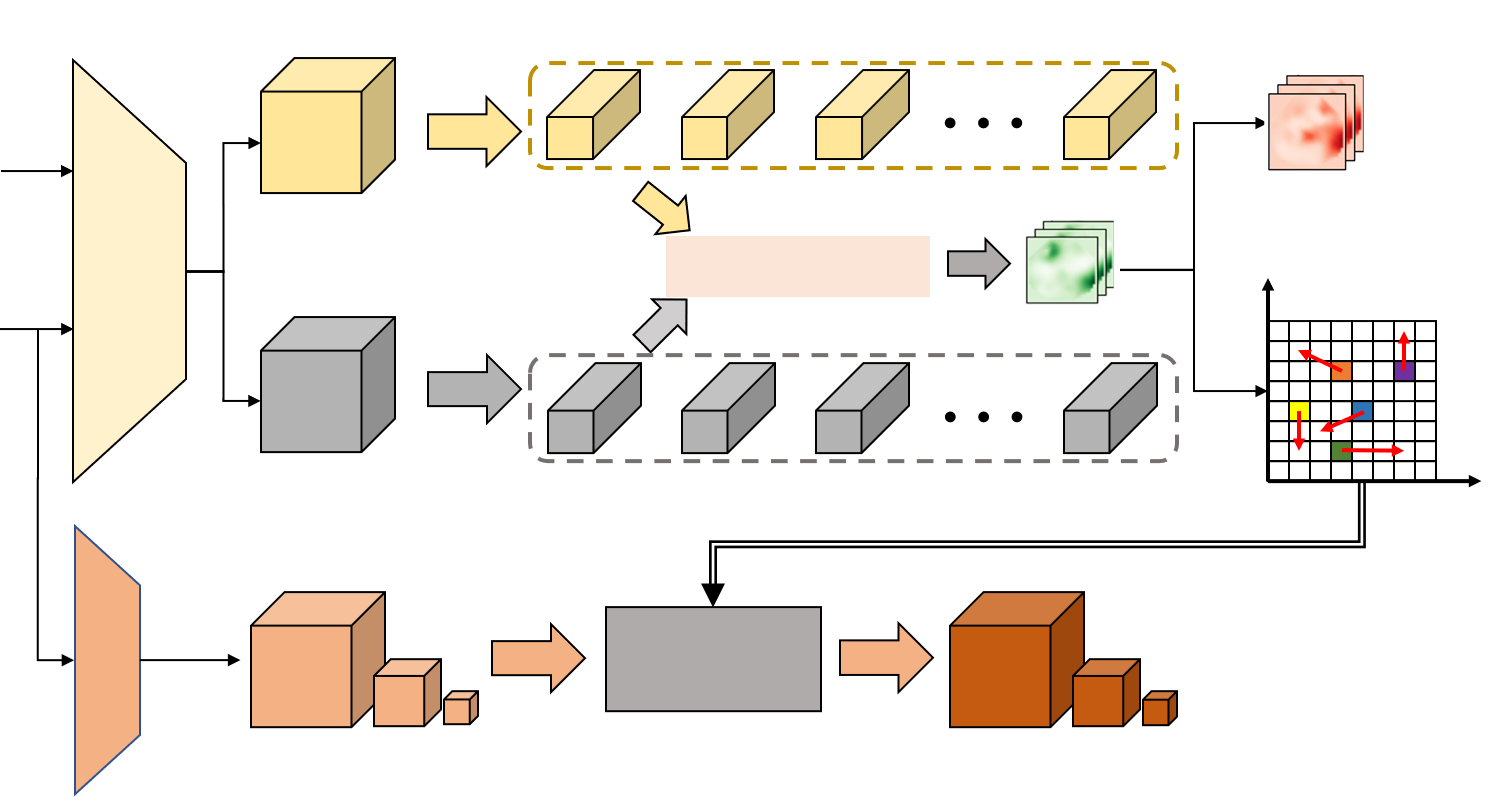}
\put(17.5,18.5){\color{black}{\footnotesize $F_{\textit{LR}}^{\textit{tex}}$}}
\put(17.5,36){\color{black}{\footnotesize $F_{\textit{Ref}}^{\textit{tex}}$}}
\put(20,0){\color{black}{\footnotesize $F_{\textit{Ref}}^{i}$}}
\put(46,8){\color{black}{\footnotesize ${\cal W}$}}
\put(67,0){\color{black}{\footnotesize ${\tilde F_{\textit{Ref}}^i}$}}
\put(36.9,39.5){\color{black}{\footnotesize $Q_1$}}
\put(46.3,39.5){\color{black}{\footnotesize $Q_2$}}
\put(55.3,39.5){\color{black}{\footnotesize $Q_3$}}
\put(70.3,39.6){\color{black}{\footnotesize $Q_{l1}$}}
\put(82,50){\color{black}{\footnotesize Pre-offset}}
\put(87,38){\color{black}{\footnotesize flow}}
\put(83,35){\color{black}{\footnotesize information}}
\put(-1,33){\color{black}{\footnotesize $I_{\textit{LR}}$}}
\put(-2,43.5){\color{black}{\footnotesize $I_{\textit{Ref}}$}}

\put(44.3,34.6){\color{black}{\footnotesize Inner Product}}

\put(36.9,19.8){\color{black}{\footnotesize $K_1$}}
\put(46.3,19.8){\color{black}{\footnotesize $K_2$}}
\put(55.3,19.8){\color{black}{\footnotesize $K_3$}}
\put(70.3,19.8){\color{black}{\footnotesize $K_{l2}$}}

\put(70,34.3){\color{black}{\footnotesize ${\cal P}$}}
\begin{turn}{270}
\put(-42,9.5){\color{black}{\footnotesize Pretrained}}
\put(-45.4,6.5){\color{black}{\footnotesize Texture Encoder}}
\put(-14.4,6.0){\color{black}{\footnotesize VGG19}}
\end{turn}
\end{overpic}
\end{center}%\vspace{-0.1cm}
\caption{The architecture of the correlation-based texture warp (CTW).}
\label{fig:ctw}
%\vspace{-0.1cm}
\end{figure}
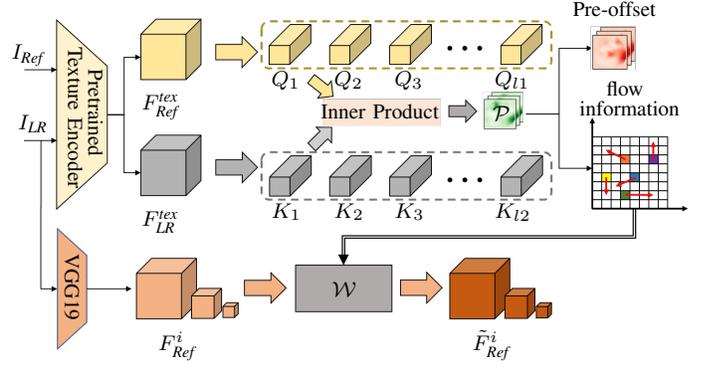
\begin{gather}
\label{deqn_ex1a}
\hat F_{\textit{Ref}}^{\textit{tex}},\hat F_{\textit{LR}}^{\textit{tex}} = \textit{unfold}\left( {F_{\textit{Ref}}^{\textit{tex}},F_{\textit{LR}}^{\textit{tex}}} \right),\\
{{\cal M}_{i,j}} = \left( {\hat F{{_{\textit{LR}}^{\textit{tex}}}^\mathrm{T}} \cdot \hat F_{\textit{Ref}}^{\textit{tex}}} \right) = \left\langle\frac{Q_i}{\left\|{Q_i}\right\|},\frac{K_j}{\left\|{K_j}\right\|}\right\rangle,
\end{gather}
where $\hat F_{\textit{LR}}^{\textit{tex}}$ and $\hat F_{\textit{Ref}}^{\textit{tex}}$ respectively represent the patch features of $F_{\textit{LR}}^{\textit{tex}}$ and $F_{\textit{Ref}}^{\textit{tex}}$ after being split into patches. $\frac{Q_i}{\left\|{Q_i}\right\|}$ and $\frac{K_j}{\left\|{K_j}\right\|}$ respectively represent the normalized features of the i-th patch in $\hat F_{\textit{LR}}^{\textit{tex}}$ and the j-th patch in $\hat F_{\textit{Ref}}^{\textit{tex}}$, and $\langle \cdot , \cdot \rangle$ represents the inner product operation. $\hat F{{_{\textit{LR}}^{\textit{tex}}}^\mathrm{T}}$ denotes the transpose of $\hat F{{_{\textit{LR}}^{\textit{tex}}}}$. For a given patch $Q_i$ in $\hat F_{\textit{LR}}^{\textit{tex}}$, the most similar patch $K_j$ in $\hat F_{\textit{Ref}}^{\textit{tex}}$ can be found and recorded as $P_\textit{max}^i$. The index matrix $P = \left\{ {{\cal P}_\textit{max}^1,{\cal P}_\textit{max}^2, \ldots ,{\cal P}_\textit{max}^{l1}} \right\} \in \mathbb{R}^{l1}$ is formed by recording the indices of these most similar patches.
\begin{equation}
\label{deqn_ex1a}
{\cal P}_{max}^i = \mathop {{\rm{argmax}}}\limits_j {{\cal M}_{i,j}},
\end{equation}
where ${\cal M}_{i,j}$ represents the confidence score of patch $K_j$ corresponding to patch $Q_i$ which is most similar to it. All ${\cal P}_\textit{max}^i$ form the index matrix $P$. To use optical flow to initially warp the reference features, we need to convert the index matrix $P$ into flow information. The process is shown below:
\begin{gather}
\label{deqn_ex1a}
(\mathcal{G}_y,\mathcal{G}_x) = \mathbb{G}(W_{\textit{LR}},H_{\textit{LR}}),\\
\mathcal{F} = \left[P\bmod {W_{\textit{LR}}}; \lfloor P,{W_{\textit{LR}}} \rfloor\right]  -  \left[\mathcal{G}_x;\mathcal{G}_y \right],
\end{gather}
where $[;]$ represents the concatenation of two vectors, $\mathbb{G}\left( \cdot \right)$ represents the grid function, which generates a grid with a width of $W_{\textit{LR}}$ and a height of $H_{\textit{LR}}$; $\mathcal{G}_x$ and  $\mathcal{G}_y$ denote the coordinate values along the height and width of the grid, The symbols `mod' and `$\lfloor\cdot,\cdot\rfloor$' represent the mathematical operations of module and floor division, respectively, and $\mathcal{F}$ represents the flow information. Finally, we select three different scales of feature maps $F_{\textit{Ref}}^i$ extracted by the pre-trained VGG19~\cite{vgg19}. The reason for choosing VGG19 is that it has a strong feature extraction ability and does not require training of additional feature extraction modules. Then, the reference features $F_{\textit{Ref}}^i$ of the three corresponding scales are warped by optical flow using the flow information of different scales, and the specific process is shown below:
\begin{gather}
\label{deqn_ex1a}
(\mathcal{G}_{y_i},\mathcal{G}_{x_i}) = \mathbb{G}(W_i,H_i),\\
{x_i},{y_i} = \textit{split}\big( {\left[ \mathcal{G}_{y_i},\mathcal{G}_{x_i} \right] - \mathcal{F}} \big),\\
\mathcal{G}_i = \left[ {\frac{{2 \times {x_i}}}{{{\rm{max}}\left( {{W_i} - 1,1} \right)}} - 1;\frac{{2 \times {y_i}}}{{{\rm{max}}\left( {{H_i} - 1,1} \right)}} - 1} \right],\\
\tilde F_{\textit{Ref}}^i = {\cal W}\left( {F_{\textit{Ref}}^i,\mathcal{G}_i} \right),
\end{gather}
where $H_i$ and $W_i$ respectively represent the height and width of $F_{\textit{Ref}}^i$, $\textit{split}\left( \cdot \right)$ represents the separation of two vectors according to the concatenated channels, $x_i, y_i \in \mathbb{R}^{H_i\times W_i}$, and ${\cal W}(\cdot,\cdot)$ represents the optical flow warping function.
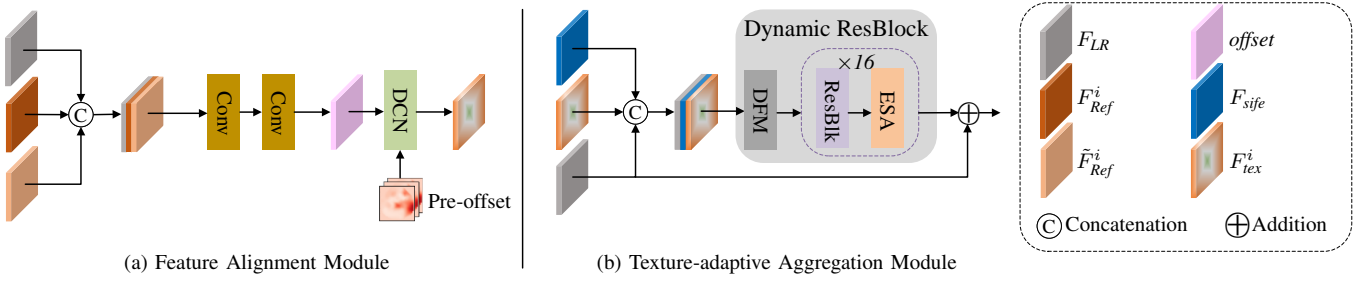
\begin{figure*}[!tbp]
\begin{center}
\begin{overpic}[width=0.99\textwidth]{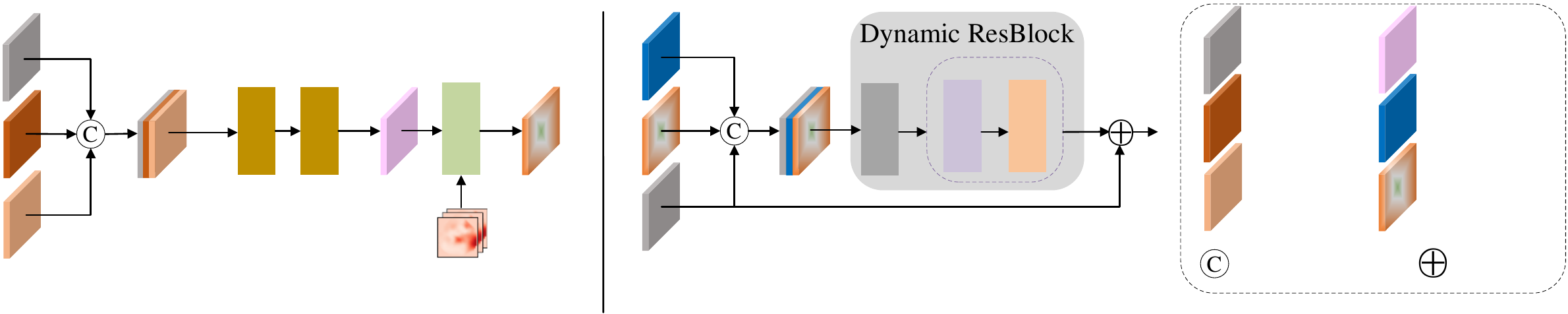}
\put(79.5,16.5){\color{black}{\footnotesize $F_\textit{LR}$}}
\put(61.7,14.8){\color{black}{\footnotesize \textit{$\times$16}}}
\put(79.5,12.2){\color{black}{\footnotesize $F_\textit{Ref}^i$}}
\put(79.5,7.5){\color{black}{\footnotesize $\tilde{F}_\textit{Ref}^i$}}
\put(90.8,16.5){\color{black}{\footnotesize $\textit{offset}$}}
\put(90.8,12.2){\color{black}{\footnotesize $F_\textit{sife}$}}
\put(90.8,7.5){\color{black}{\footnotesize $F_\textit{tex}^i$}}
\put(31.5,4.5){\color{black}{\footnotesize Pre-offset}}
\put(78.7,2.8){\color{black}{\footnotesize Concatenation}}
\put(92.5,2.8){\color{black}{\footnotesize Addition}}
\put(9,0){\color{black}{\footnotesize (a) Feature Alignment Module}}
\put(44,0){\color{black}{\footnotesize (b) Texture-adaptive Aggregation Module}}
\begin{turn}{270}
\put(-13.2,15.9){\color{black}{\footnotesize Conv}}
\put(-13.2,19.9){\color{black}{\footnotesize Conv}}
\put(-13.4,28.9){\color{black}{\footnotesize DCN}}
\put(-13.4,55.6){\color{black}{\footnotesize DFM}}
\put(-14,60.9){\color{black}{\footnotesize ResBlk}}
\put(-13.2,65){\color{black}{\footnotesize ESA}}

\end{turn}
\end{overpic}
\end{center}%\vspace{-0.1cm}
\caption{The structure of the feature alignment module (FAM) and texture-adaptive aggregation module (TAAM). The gray background in TAAM signifies the Dynamic ResBlk (DRB) that was designed by us.}
\label{fig:FAM&TAAM}
%\vspace{-0.1cm}
\end{figure*}
\subsection{Multi-scale Dynamic Texture Aggregation}
Using an effective texture transfer based on a corresponding matching relationship is another important goal of the RefSR task. To more effectively transfer and aggregate the textures in reference images, we propose a multi-scale dynamic texture transfer module based on the U-Net network~\cite{unet}, as shown in the gray background in Fig.1. Using the multi-scale characteristics of the U-Net network, we can progressively aggregate the texture features in multi-scale reference images and learn to generate richer textures. Unlike the direct texture transfer methods used in ~\cite{ttsr,srntt}, we use specific deformable convolutions~\cite{dcn,dcnv2} for texture alignment between $F_{\textit{LR}}$ and $F_{\textit{Ref}}^i$ for RefSR tasks, and finally use a texture-adaptive aggregation module to complete texture transfer and aggregation. In addition to this main task of texture transfer, RefSR often struggles to reconstruct high-frequency details from the LR itself. To address this issue, we use the SISR method to reconstruct LR's own features and embed the reconstructed feature $F_{\textit{sife}}$ into the feature aggregation process. In this way, not only can we supplement the detailed textures that are difficult to reconstruct in RefSR, but also limit the introduction of irrelevant textures to some extent.
\begin{equation}
\label{deqn_ex1a}
F_{\textit{sife}} = \textit{SIFE}(I_{\textit{LR}}).
\end{equation}
We chose the same SISR baseline used in~\cite{tade} to ensure a more equitable comparison. Nevertheless, we removed the last upsampling stage which is present in SISR.
\subsubsection{Texture Alignment Module}
To more accurately transfer the texture features in the multi-scale reference feature $F_{\textit{Ref}}^i$, we use specific deformable convolutions designed for RefSR to achieve accurate texture alignment. As shown in the flowchart in Fig.3 (a), to obtain the offset required for deformable convolution, we concatenate $F_{\textit{LR}}$, $F_{\textit{Ref}}^i$, and $\tilde F_{\textit{Ref}}^i$ to obtain the offset $\Delta P_k$. This is because using the optically distorted reference feature to guide deformable convolution training can make the training process more stable.
\begin{equation}
\label{deqn_ex1a}
{{\Delta \cal P}_k}=\textit{Conv}\Bigg( {\textit{Conv}\bigg( {\Big[ {{F_{\textit{LR}}};F_{\textit{Ref}}^i;\tilde F_{\textit{Ref}}^i} \Big]} \bigg)} \Bigg),
\end{equation}
where $\Delta {\cal P}_k$ represents the offset, and $\textit{Conv}\left( \cdot \right)$ represents the convolution layer. After this, for each patch ${\cal P}_{\textit{LR}}$ in LR, we used the previously obtained index matrix P to find the corresponding most similar patch ${\cal P}_{\textit{Ref}}$ in $F_{\textit{Ref}}^i$. We use $\Delta {\cal P}$ to represent the spatial difference between ${\cal P}_{\textit{LR}}$ and ${\cal P}_{\textit{Ref}}$, that is, ${\Delta\cal P} = {{\cal P}_{\textit{LR}}} - {{\cal P}_{\textit{Ref}}}$, which is the pre-offset output by CTW. Finally, the improved deformable convolution is used to aggregate ${\cal P}_{\textit{LR}}$ and its surrounding textures. The specific process is shown below.
\begin{equation}
\label{deqn_ex1a}
F_{\textit{tex}}^p = \mathop \sum \limits_{k = 1}^K {w_k}\cdot y\left( {{{\cal P}_{\textit{LR}}} + {\Delta \cal {P}} + {{\cal P}_k} + {{\Delta \cal {P}}_k}} \right)\cdot {\Delta {m}_k},
\end{equation}
where $y$ represents the original reference feature, ${{\cal P}_k} \in \left\{ {\left( { - 1,{\rm{\;}}1} \right),{\rm{\;}}\left( { - 1,{\rm{\;}}0} \right),\ldots ,\left( {1,{\rm{\;}}1} \right)} \right\}$; $\Delta {P}_k$ represents the learnable offset; $w_k$ represents the convolution weight; $\Delta{m}_k$ represents the modulation scalar; $F_{\textit{tex}}^p$ represents the reference feature after alignment at position $p$. Through the above texture alignment method based on deformable convolution, the surrounding textures of the most similar patches in each corresponding reference feature can be aggregated, fully utilizing the contextual information in each patch, thus providing a guarantee for subsequent texture transfer.
\begin{figure*}[!tbp]
\begin{center}
\begin{overpic}[width=0.8\textwidth]{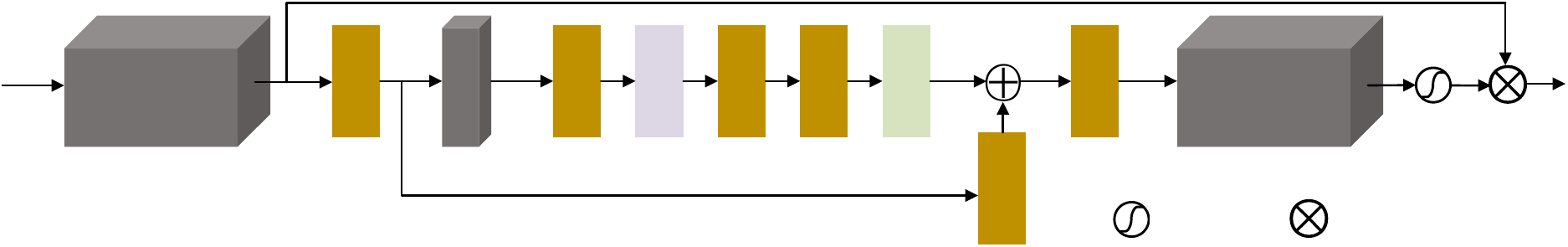}
\put(4,4.5){\color{black}{\footnotesize$H \times W \times C$}}
\put(26,4.5){\color{black}{\footnotesize $H\times W\times \frac{C}{4}$}}
\put(75,4.5){\color{black}{\footnotesize $H\times W\times C$}}
\put(74,1.2){\color{black}{\footnotesize Sigmoid}}
\put(85.5,1.2){\color{black}{\footnotesize Element-wise product}}
\begin{turn}{270}
\put(-12.5,22.1){\color{black}{\footnotesize Conv}}
\put(-12.5,36.2){\color{black}{\footnotesize Conv}}
\put(-13.5,41.5){\color{black}{\footnotesize Pooling}}
\put(-12.5,46.7){\color{black}{\footnotesize Conv}}
\put(-12.5,52){\color{black}{\footnotesize Conv}}
\put(-13.1,57.2){\color{black}{\footnotesize Biliear}}
\put(-12.5,69.3){\color{black}{\footnotesize Conv}}
\put(-5.5,63.3){\color{black}{\footnotesize Conv}}
\end{turn}
\end{overpic}
\end{center}%\vspace{-0.1cm}
\caption{The structure of Enhanced Spatial Attention (ESA).}
\label{fig:esa}
%\vspace{-0.1cm}
\end{figure*}

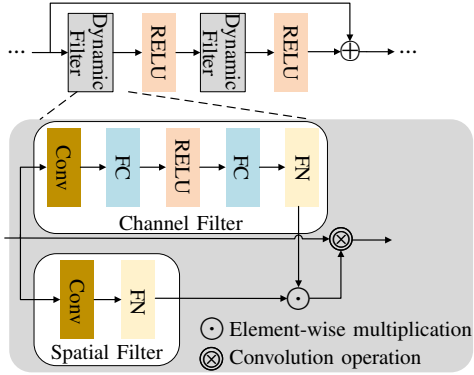
\begin{figure}[!tbp]
\begin{center}
\begin{overpic}[width=0.345\textwidth]{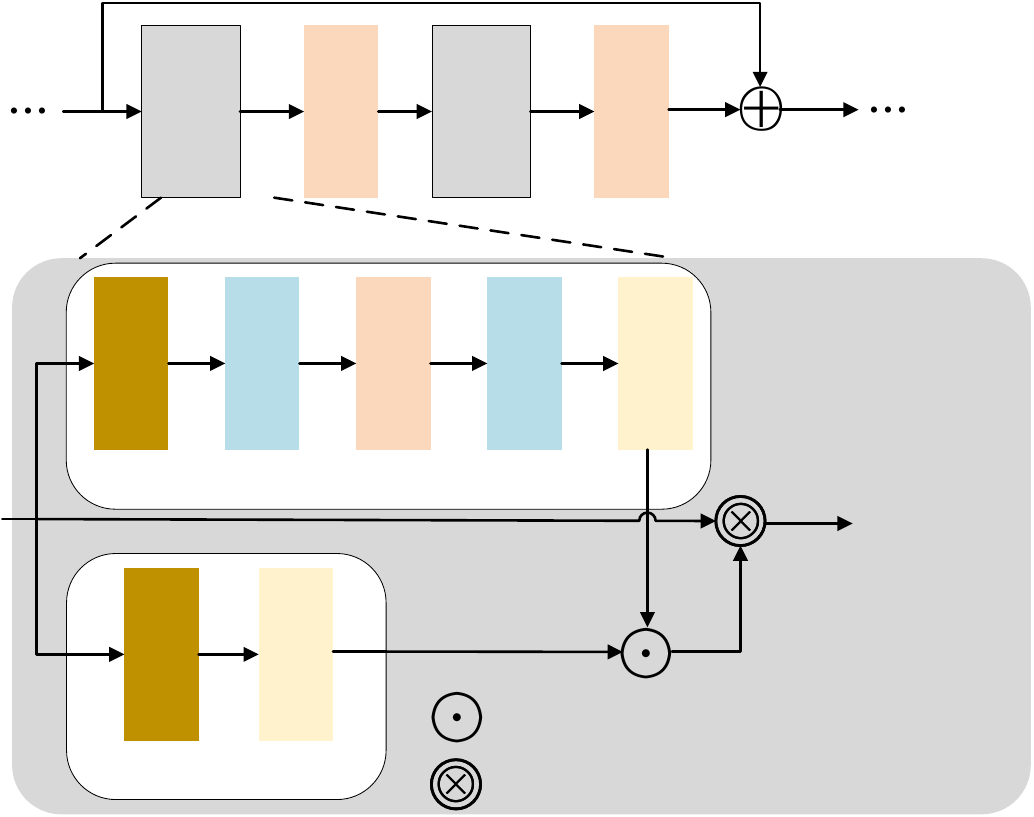}
\put(24.5,30.5){\color{black}{\footnotesize Channel Filter}}
\put(10,3){\color{black}{\footnotesize Spatial Filter}}
\put(48,2){\color{black}{\footnotesize Convolution operation}}
\put(48,8){\color{black}{\footnotesize Element-wise multiplication}}
\begin{turn}{270}
\put(-76.6,19){\color{black}{\footnotesize Dynamic}}
\put(-73,14.2){\color{black}{\footnotesize Filter}}
\put(-74,31.4){\color{black}{\footnotesize RELU}}
\put(-76.6,47.2){\color{black}{\footnotesize Dynamic}}
\put(-73,42.4){\color{black}{\footnotesize Filter}}
\put(-74,59.5){\color{black}{\footnotesize RELU}}

\put(-48.5,11.2){\color{black}{\footnotesize Conv}}
\put(-46.5,23.8){\color{black}{\footnotesize FC}}
\put(-49.8,36.5){\color{black}{\footnotesize RELU}}
\put(-46.5,49){\color{black}{\footnotesize FC}}
\put(-46.5,62){\color{black}{\footnotesize FN}}

\put(-20,14.1){\color{black}{\footnotesize Conv}}
\put(-18,27){\color{black}{\footnotesize FN}}
\end{turn}
\end{overpic}
\end{center}%\vspace{-0.1cm}
\caption{The structure of dynamic filter module (DFM). `FC' denotes the fully connected layer and `GAP' denotes the global average pooling. `FN' denotes filter normalization proposed in~\cite{dedyfilter}.}
\label{fig:DFM}
%\vspace{-0.1cm}
\end{figure}
\subsubsection{Texture-Adaptive Aggregation Module}
To effectively aggregate the features of $F_{\textit{LR}}$, $F_{\textit{tex}}^i$, and $F_{\textit{sife}}$. Inspired by~\cite{dedyfilter}, we propose TAAM for self-adapting transfer and aggregating related texture features, as shown in the flowchart in Fig.3 (b). Specifically, we concatenate the aligned texture feature $F_{\textit{tex}}^i$ with $F_{\textit{LR}}$ and $F_{\textit{sife}}$, and input them into a convolution layer. Then, we use the dynamic residual block to transfer and aggregate the related textures in the reference feature to obtain the output $F_{\textit{agg}}$. It is worth noting that we only embed $F_{\textit{sife}}$ in the TAAM module corresponding to the smallest scale, that is, only the feature mapping of the smallest scale is used. The other TAAM modules at other scales only aggregate $F_{\textit{LR}}$ and $F_{\textit{tex}}$ features.
\begin{equation}
\label{deqn_ex1a}
{F_{\textit{agg}}^{\textit{rec}}} = \textit{DRB}\Bigg({\textit{Conv}\bigg({\Big[ {{F_{\textit{tex}}};{F_{\textit{LR}}};{F_{\textit{sife}}}} \Big]} \bigg )} \Bigg ) + {F_{\textit{LR}}}.
\end{equation}
To train the second model, we reused the feature map $F_{\textit{agg}}^{\textit{rec}}$ created by the first model. As a result, we added this feature map to the feature aggregation process to enhance the texture features. Equation 17 can be expressed in the following form:
\begin{equation}
\label{deqn_ex1a}
{F_{\textit{agg}}^{all}} = \textit{DRB}\Bigg({\textit{Conv}\bigg({\Big[ {{F_{\textit{tex}}};{F_{\textit{LR}}};{F_{\textit{sife}}}}; F_{\textit{agg}}^{\textit{rec}}\Big]} \bigg )} \Bigg ) + {F_{\textit{LR}}}.
\end{equation}
The DRB module consists of two decoupled dynamic filters and a ResBlock with an ESA (Enhanced Spatial Attention)~\cite{esa}. The decoupled dynamic filter and ESA are shown in Fig.5 and Fig.4. The decoupled dynamic filter is inspired by~\cite{dedyfilter} and is decoupled into channel filters and spatial filters, which can effectively perceive the related texture content between $F_{\textit{LR}}$ and $F_{\textit{tex}}^i$, $F_{\textit{sife}}$. The decoupled dynamic filter operation can be written as:
\begin{equation}
\label{deqn_ex1a}
{\hat F_{\left( {k,i} \right)}} = \mathop \sum \limits_{j \in \Omega \left( i \right)} H_i^\textit{sf}( {{p_i}\!-\!{p_j}} )H_k^\textit{cf}( {{p_i}\!- \!{p_j}} ){F_{\left( {k,j} \right)}},
\end{equation}
where $\Omega(i)$ represents the convolution window around the i-th pixel, $p_i$ and $p_j$ represent the pixel coordinates, $H_i^\textit{sf}(\cdot)$ represents the spatial filter, $H_i^\textit{cf}(\cdot)$ represents the channel filter, and the feature value of the $k$-th channel and $j$-th pixel before and after dynamic filtering are denoted as $F_{\left( {k,i} \right)}$ and $\hat F_{\left( {k,i} \right)}$, respectively. Then the routing weight and the final aggregated features can be generated:
\begin{gather}
\label{deqn_ex1a}
(\omega_1, \omega_2, \ldots, \omega_n) = \Big(\gamma^\textit{sf} \big(\hat H_{i}^\textit{sf}\big)+ \beta^\textit{sf}\Big) \odot \Big(\gamma^\textit{cf} \big(\hat H_{i}^\textit{cf}\big) + \beta^\textit{cf}\Big), \\
F_{\textit{tex}}^{'} = (\omega_1, \omega_2, ,\ldots, \omega_n) * F_{\textit{tex}},
\end{gather}
where, $\hat H_{i}^\textit{sf}$ and $\hat H_{i}^\textit{cf}$ represent the values obtained from the spatial and channel filter branches, respectively, after normalization is applied, while $\omega$ denotes the routing weights. $\gamma^\textit{sf}$, $\gamma^\textit{cf}$, $\beta^\textit{sf}$, and $\beta^\textit{cf}$ are similar to BN~\cite{bn} and specify the learnable mean and standard deviation of the two branches. `$\odot$' and `$*$' are used to denote element-wise multiplication and the convolutional operation, respectively.

ESA has been proven to be efficient and effective in previous work~\cite{esa,bsrn}. This is because it uses $1\times1$ convolution and $3\times3$ convolution with a stride of 2 to compress the channel size and spatial size respectively, and further reduces the feature size using max pooling. The specific process of ESA is as follows:
\begin{gather}
\label{deqn_ex1a}
{F_0} = \textit{Conv}_1^1\left( F \right), \\
{F_1} = \textit{Conv}_1^1\left( {F_0} \right), \\
{F_2} = \mathcal{B}\Bigg( {\textit{Conv}_1^3\bigg( {\textit{Conv}_1^3\Big( {\textit{Pool}\big( {\textit{Conv}_2^3( {{F_0}} )} \big)} \Big)} \bigg)} \Bigg), \\
F' = \sigma \Big( {\textit{Conv}_1^1({F_1} + {F_2}} )\Big) \otimes F,
\end{gather}
where $\textit{Conv}_a^b(\cdot)$ represents a convolution layer with kernel size a and stride b, $\textit{Pool}(\cdot)$ represents a max pooling layer, $\mathcal{B}(\cdot)$ represents bilinear interpolation, $\sigma(\cdot)$ represents the sigmoid function, `$\otimes$' represents element-wise product, $F$ and $F^{'}$ represent input feature and output feature respectively.

ResBlock with the ESA module can enhance the related texture features of $F_{\textit{LR}}$, and aggregate reference features with high relevance while suppressing interference features with low relevance. It is worth noting that this attention module is lightweight and only adds a small number of parameters.

This attention-based texture-adaptive aggregation method; not only transfers and fuses effective textures from reference images and reduces interference from irrelevant textures, it also ensures that the features $F_{\textit{sife}}$ reconstructed by the SISR method are well integrated into $F_{\textit{LR}}$. By aggregating $F_{\textit{sife}}$ features, it not only makes up for the defect that reference-based super-resolution is difficult to reconstruct its own texture, it also suppresses the generation of irrelevant textures to a large extent.

\subsection{Loss Functions}
\paragraph{Reconstruction loss} To ensure the model has an excellent texture transfer ability and image reconstruction ability, we use the following reconstruction loss to train the model.
\begin{equation}
\label{deqn_ex1a}
{{\cal L}^{\textit{rec}}} = \;\left\| I_\textit{HR} - I_{\textit{SR}} \right\|_1,
\end{equation}
where $I_{HR}$ represents the ground truth image, $I_{\textit{SR}}$ represents the super-resolved image. ${\left\|\cdot\right\|}_1$ represents $l_1$ norm. Only using reconstruction loss to train the model will cause the image to be too smooth.
\paragraph{Perceptual loss} By calculating perceptual loss~\cite{preloss} in the feature domain, the generated image can be more semantically similar to GT. Perceptual loss is shown as follows:
\begin{equation}
\label{deqn_ex1a}
{{\cal L}^{\textit{per}}} = \frac{1}{V}\mathop \sum \limits_{i = 1}^C \left\|{\phi _i}\left( {{I_\textit{HR}}} \right) - {\phi _i}\left( {{I_{\textit{SR}}}} \right) \right\|_F,
\end{equation}
where $\phi _i(\cdot)$ represents the $i$-th intermediate layer of VGG19~\cite{vgg19}. ${\left\|\cdot\right\|}_F$ represents Frobenius norm, $C$ and $V$ represent the number of channels and volume of feature maps respectively.
\paragraph{Adversarial loss} The generator $G$ and discriminator $\cal D$ improve together in a game against each other, ensuring the model is able to generate output images with pleasing visual effects. The adversarial loss we choose is WGAN~\cite{wgan}, which is shown as follows:
\begin{equation}
\label{deqn_ex1a}
{{\cal L}^{\textit{adv}}} =  - {\cal D}\left( {{I_{\textit{SR}}}} \right).
\end{equation}
During the training process, the loss of discriminator $\cal D$ is shown as follows:
\begin{equation}
\label{deqn_ex1a}
{{\cal L}^{\cal D}} = {\cal D}\left( {{I_{\textit{SR}}}} \right) - {\cal D}\left( {{I_{GT}}} \right) + \lambda {\left( \left\|{{\nabla _{\hat I}}{\cal D}{{\left( {\hat I} \right)}}}\right\|_2 - 1 \right)}^2,
\end{equation}
where $\nabla _{\hat I}$ represents the random convex combination of $I_{HR}$ and $I_{\textit{SR}}$.

Finally, the total loss function is shown as follows:
\begin{equation}
\label{deqn_ex1a}
{{\cal L}^\textit{all}} = {\lambda _1}{{\cal L}^{\textit{rec}}} + {\lambda _2}{{\cal L}^{\textit{per}}} + {\lambda _3}{{\cal L}^{\textit{adv}}},
\end{equation}
where $\lambda_1$, $\lambda_2$, and $\lambda_3$ are respectively the weight coefficients for each loss.
\section{Experiments}
This section commences by presenting the datasets essential to the training and testing of the models. Subsequently, we comparatively analyze several super-resolution methods along various aspects for our approach. Ablation studies are conducted on the SIFE and DRB components, along with the feature reuse framework. Lastly, we evaluate the efficacy of our proposed approach against other super-resolution methods in a practical implementation.
\begin{table*}[!t]
\caption{PSNR/SSIM are the evaluation metrics uesd to compare the other methods quantitatively. The model is trained using only reconstruction loss (-$rec$). Text highlighted in bold indicates the most favorable outcome.\label{tab:table1}}
\centering
\renewcommand\arraystretch{1.5}
\begin{tabular}{c|c|c|c|c|c|c}
\hline
\multirow{2}{*}{SR paradigms} & \multirow{2}{*}{Method} & CUFED5~\cite{srntt} & Sun80~\cite{sun80} & Urban100~\cite{urban100} & Manga109~\cite{manga} & WR-SR~\cite{c2} \\ \cline{3-7} & &
PSNR/SSIM & PSNR/SSIM & PSNR/SSIM & PSNR/SSIM & PSNR/SSIM \\ \hline\hline
\multicolumn{1}{c|}{\multirow{5}{*}{SISR}} & \multicolumn{1}{c|}{SRCNN~\cite{srcnn}}
& 25.33/0.745 & 28.26/0.781 & 24.41/0.738 & 27.12/0.850 & 27.27/0.767 \\ \cline{2-7}
\multicolumn{1}{c|}{}                                           & EDSR~\cite{edsr}
& 25.93/0.777 & 28.52/0.792 & 25.51/0.783 & 28.93/0.891 & 28.07/0.793 \\ \cline{2-7}
\multicolumn{1}{c|}{}                                           & ENet~\cite{enet}
& 24.24/0.695 & 26.24/0.702 & 23.63/0.711 & 25.25/0.802 & 25.47/0.699 \\ \cline{2-7}
\multicolumn{1}{c|}{}                                           & RCAN~\cite{rcan}
& 26.33/0.781 & 29.97/0.814 & 25.99/0.787 & 30.11/0.908 & 27.91/0.793 \\ \cline{2-7}
\multicolumn{1}{c|}{}                                           & RRDB~\cite{esrgan}
&26.41/0.783 & 29.99/0.814 & 25.98/0.788 & 29.87/0.907 & 27.96/0.793  \\ \hline \hline
\multirow{11}{*}{RefSR} & Cross-Net~\cite{crossnet}
& 25.48/0.764 & 28.52/0.793 & 25.11/0.764 & 23.36/0.741 & -  \\ \cline{2-7}
                        & SSEN-$rec$~\cite{ssen}
& 26.78/0.791 & - & - & - & - \\ \cline{2-7}
                        & SRNTT-$rec$~\cite{srntt}
& 26.24/0.784 & 28.54/0.793 & 25.50/0.783 & 28.95/0.885 & 27.59/0.780 \\ \cline{2-7}
                        & TTSR-$rec$~\cite{ttsr}
& 27.09/0.804 & 30.02/0.814 & 25.87/0.784 & 30.09/0.907 & 27.97/0.792 \\ \cline{2-7}
                        & MASA-$rec$~\cite{masa}
& 27.54/0.814 & 30.15/0.815 & 26.09/0.786 & 30.28/0.909 & 28.19/0.796 \\ \cline{2-7}
                        & $C^2$-Matching-$rec$~\cite{c2}
& 28.24/0.841 & 30.18/0.817 & 26.03/0.785 & 30.47/0.911 & 28.32/0.801  \\ \cline{2-7}
                        & WTRN-$rec$\cite{wtrn}
& 27.33/0.810 & 30.11/0.816 & 26.00/0.787 & 30.37/0.909 & -        \\ \cline{2-7}
                       & TADE-$rec$~\cite{tade}
& 28.64/0.850 & 30.31/0.820 & 26.71/0.807 & \bf{31.23/0.917} & 28.52/0.807 \\ \cline{2-7}
                       & DATSR-$rec$~\cite{datsr}
& 28.72/0.856 & 30.20/0.818 & 26.52/0.798 & 30.49/0.912 & 28.34/0.805 \\ \cline{2-7}
                        & RRSR-$rec$~\cite{rrsr}
& 28.83/0.856 & 30.13/0.816 & 26.21/0.790 & 30.91/0.913 & 28.41/0.804 \\ \cline{2-7}
                        & FRFSR-$rec$ (Ours)
& \bf{29.18/0.865} & \bf{30.35/0.822} & \bf{26.84/0.811} & 31.15/\bf{0.917} & \bf{28.67/0.811} \\ \hline
\end{tabular}
\end{table*}

\begin{table*}[!t]
\caption{The model is trained using all losses and the results are compared with PSNR/SSIM. Text highlighted in bold indicates the most favorable outcome.\label{tab:table2}}
\centering
\renewcommand\arraystretch{1.5}
\begin{tabular}{c|c|c|c|c|c|c}
\hline
\multirow{2}{*}{SR paradigms} & \multirow{2}{*}{Method}
& CUFED5~\cite{srntt} & Sun80~\cite{sun80} & Urban100~\cite{urban100} & Manga109~\cite{manga} & WR-SR~\cite{c2}     \\ \cline{3-7}
& & PSNR/SSIM & PSNR/SSIM & PSNR/SSIM & PSNR/SSIM & PSNR/SSIM \\ \hline\hline
\multicolumn{1}{c|}{\multirow{3}{*}{SISR}} & \multicolumn{1}{c|}{SRGAN~\cite{srgan}}
& 24.40/0.702 & 26.76/0.725 & 24.07/0.729 & 25.12/0.802 & 26.21/0.728 \\ \cline{2-7}
\multicolumn{1}{c|}{}                      & ESRGAN~\cite{esrgan}
& 21.90/0.633 & 24.18/0.651 & 20.91/0.620 & 23.53/0.797 & 26.07/0.726 \\ \cline{2-7}
\multicolumn{1}{c|}{}                      & RankSRGAN~\cite{rankgan}
& 22.31/0.635 & 25.60/0.667 & 21.47/0.624 & 25.04/0.803 & 26.15/0.719 \\ \cline{2-7}
\hline
\multirow{10}{*}{RefSR}                    & SRNTT~\cite{srntt}
& 25.61/0.764 & 27.59/0.756 & 25.09/0.774 & 27.54/0.862 & 26.53/0.745 \\ \cline{2-7}
                                           & TTSR~\cite{ttsr}
& 25.53/0.765 & 28.59/0.774 & 24.62/0.747 & 28.70/0.886 & 26.83/0.762 \\ \cline{2-7}
                                            & MASA~\cite{masa}
& 24.92/0.729 & 27.12/0.708 & 23.78/0.712 & 27.34/0.848 & 25.76/0.717 \\ \cline{2-7}
                                            & $C^2$-Matching~\cite{c2}
& 27.16/0.805 & 29.75/0.799 & 25.52/0.764 & 29.73/0.893 & 27.80/0.780 \\ \cline{2-7}
                                            & WTRN~\cite{wtrn}
&25.98/0.761 & 28.46/0.756 & 24.88/0.747 &29.18/0.878  & -            \\ \cline{2-7}
                                            & TADE~\cite{tade}
& 27.37/0.816 & 28.85/0.768 & 25.80/0.776 & 30.12/0.889 & 27.40/0.769 \\ \cline{2-7}
                                            & DATSR~\cite{datsr}
& 27.95/0.835 & 29.77/0.800 & 25.92/0.775 & 29.75/0.893 & 27.87/0.787 \\ \cline{2-7}
                                            & RRSR~\cite{rrsr}
& 28.09/0.835 & 29.57/0.793 & 25.68/0.767 & 29.82/0.893 & 27.89/0.784 \\ \cline{2-7}
                                            & FRFSR(Ours)
& \bf{28.71/0.852} & \bf{29.89/0.804} & \bf{26.65/0.802} & \bf{30.89/0.906} & \bf{28.27/0.793} \\ \hline
\end{tabular}
\end{table*}
\subsection{Datasets and Metrics}
\paragraph{Training Dataset}
We use CUFED~\cite{srntt} to train our model, which consists of two parts: a training set and a testing set. The training set contains 11871 pairs of input and reference images, each with a resolution of 160$\times$160.
\paragraph{Testing Dataset}
Our study evaluates the efficiency of our model across five benchmark datasets: CUFED5~\cite{srntt}, Urban100~\cite{urban100}, Manga109~\cite{manga}, Sun80~\cite{sun80}, and MR-SR~\cite{c2}. CUFED5 consists of 126 image pairs, each with an input image and five distinct reference images. Urban100 contains 100 images of urban buildings, from which we use the LR image as the reference due to its strong self-similarity. For Manga109, we randomly selected a single reference image from a total of 109 anime images. Sun80 is composed of 80 input images, each with 20 reference images, and one of them was randomly selected as the reference. MR-SR is similar to CUFED5, but with a one-to-one correspondence between the LR and Ref images, resulting in a total of 80 image pairs. Our metrics for evaluation consisted of PSNR and SSIM calculated on the Y channel in the YCbCr color space.
\paragraph{Implementation Details}
To obtain the LR inputs, we downsample the HR images by a scale factor of 4. For data augmentation, we apply horizontal flip, vertical flip, and random rotation. To increase the training difficulty and improve the performance of long-distance feature alignment, we divide the reference images into patches and shuffle them randomly. We use the official RRDB~\cite{esrgan} parameters as the pre-trained model for the single image feature embedding module, which we train in two stages. First, we use ${{\cal L}^{\textit{rec}}}$ as the only loss function. Second, we use ${{\cal L}^{\textit{rec}}}$, ${{\cal L}^{\textit{per}}}$, and ${{\cal L}^{\textit{adv}}}$ for joint supervision. During the training process, we choose the Adam optimizer and set the $\beta_1$ and $\beta_2$ parameters to 0.99 and 0.999, respectively. We set the initial learning rate of the model to 1e-4 and the batch size to 9. The weights ${\lambda _1}$, ${\lambda _2}$, and ${\lambda _3}$ of ${{\cal L}^{\textit{rec}}}$, ${{\cal L}^{\textit{per}}}$, and ${{\cal L}^{\textit{adv}}}$ are set to 1.0, ${10}^{-4}$, and ${10}^{-6}$, respectively.
\subsection{Comparison with State-of-the-art Methods}
We conduct quantitative and qualitative comparisons between our proposed method and some existing SISR and RefSR methods. The SISR methods are SRCNN~\cite{srcnn}, EDSR~\cite{edsr}, RCAN~\cite{rcan}, Enet~\cite{enet}, SRGAN~\cite{srgan}, ESRGAN~\cite{esrgan}, RankSRGAN~\cite{rankgan}. The RefSR methods are CrossNet~\cite{crossnet}, SSEN~\cite{ssen}, SRNTT~\cite{srntt}, TTSR~\cite{ttsr}, MASA~\cite{masa}, ${{C}^2}$-Matching~\cite{c2}, TADE~\cite{tade}, DATSR~\cite{datsr}, and RRSR~\cite{rrsr}. We train two sets of parameters, one using only the reconstruction loss (denoted by $-rec$), and the other using all losses.
\paragraph{Quantitative Comparison} As shown in Table 1, our method achieves state-of-the-art results on four benchmark datasets using only the reconstruction loss. Our method leverages effective texture matching, dynamic texture transfer, and complementary SISR features in the reconstruction process, which enables it to transfer similar textures from the high-resolution reference images in CUFED5 and WR-SR datasets to the LR images, enhancing their high-frequency information, and to transfer self-features to assist LR reconstruction on the self-similar dataset Urban100. As shown in Table 2, our model outperforms all the other methods on all datasets under the joint supervision of losses, although its performance slightly degrades compared to the results obtained when only using reconstruction loss. Interestingly, our method still maintains a significant advantage (+0.8dB) over the other RefSR methods, even with the presence of perceptual loss and adversarial loss. The quantitative comparison under the two paradigms demonstrates that our model exhibits a strong generalization ability and achieves optimal performance.
\begin{figure*}[!tbp]
\centering
\footnotesize
\renewcommand\arraystretch{1.2}
\begin{tabular}{p{0.18\textwidth}<{\centering} p{0.05\textwidth}<{\centering} p{0.049\textwidth}<{\centering} p{0.044\textwidth}<{\centering} p{0.057\textwidth}<{\centering} p{0.19\textwidth}<{\centering} p{0.049\textwidth}<{\centering} p{0.049\textwidth}<{\centering} p{0.049\textwidth}<{\centering} p{0.049\textwidth}<{\centering} }
LR        & RCAN & RRDB  & TTSR & MASA & LR & RCAN & RRDB  & TTSR & MASA \\ \hline
Reference & ~\cite{c2}   & DATSR & Ours & HR   & Reference & ~\cite{c2}   & DATSR & Ours & HR
\end{tabular}
\label{table:example}

\includegraphics[width=\textwidth]{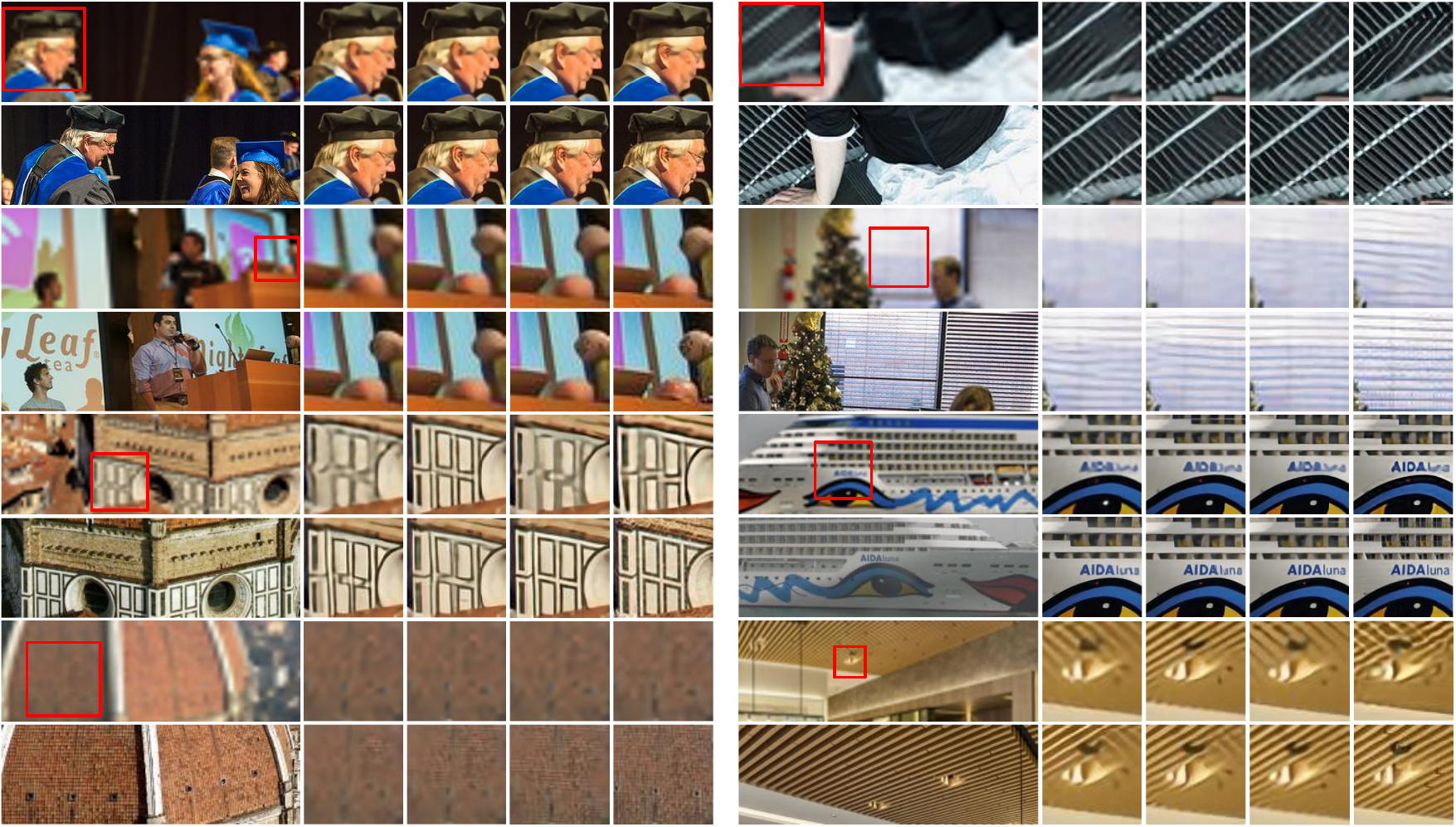}
\caption{Qualitative comparison between the model using only reconstruction loss training and the SISR and RefSR methods.}
\label{fig:results1}
\end{figure*}

\begin{figure*}[!tbp]
\centering
\footnotesize
\renewcommand\arraystretch{1.2}
\begin{tabular}{p{0.18\textwidth}<{\centering} p{0.05\textwidth}<{\centering} p{0.049\textwidth}<{\centering} p{0.044\textwidth}<{\centering} p{0.057\textwidth}<{\centering} p{0.19\textwidth}<{\centering} p{0.049\textwidth}<{\centering} p{0.049\textwidth}<{\centering} p{0.049\textwidth}<{\centering} p{0.049\textwidth}<{\centering} }
LR        & ESRGAN & SRNTT  & TTSR & MASA & LR & ESRGAN & SRNTT  & TTSR & MASA \\ \hline
Reference & ~\cite{c2}   & DATSR & Ours & HR   & Reference & ~\cite{c2}   & DATSR & Ours & HR
\end{tabular}
\label{table:example}

\includegraphics[width=\textwidth]{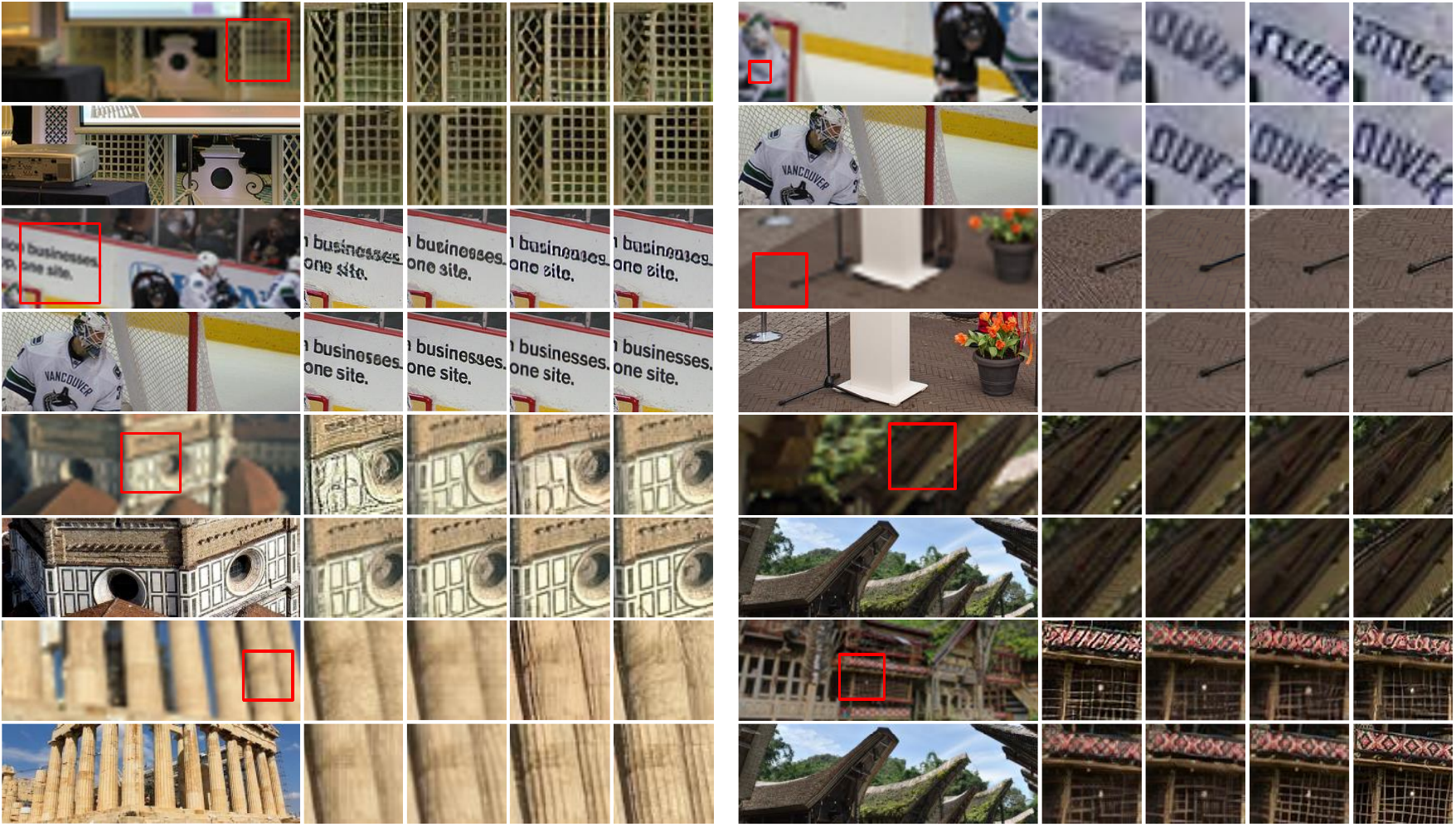}
\caption{Qualitative comparison of the model trained with all losses and the SISR and RefSR methods. It can be seen from the figure that our method can transfer and reconstruct more texture details from the reference images.}
\label{fig:results2}
\end{figure*}

\begin{figure}[!tbp]
\begin{center}
\begin{overpic}[width=0.49\textwidth]{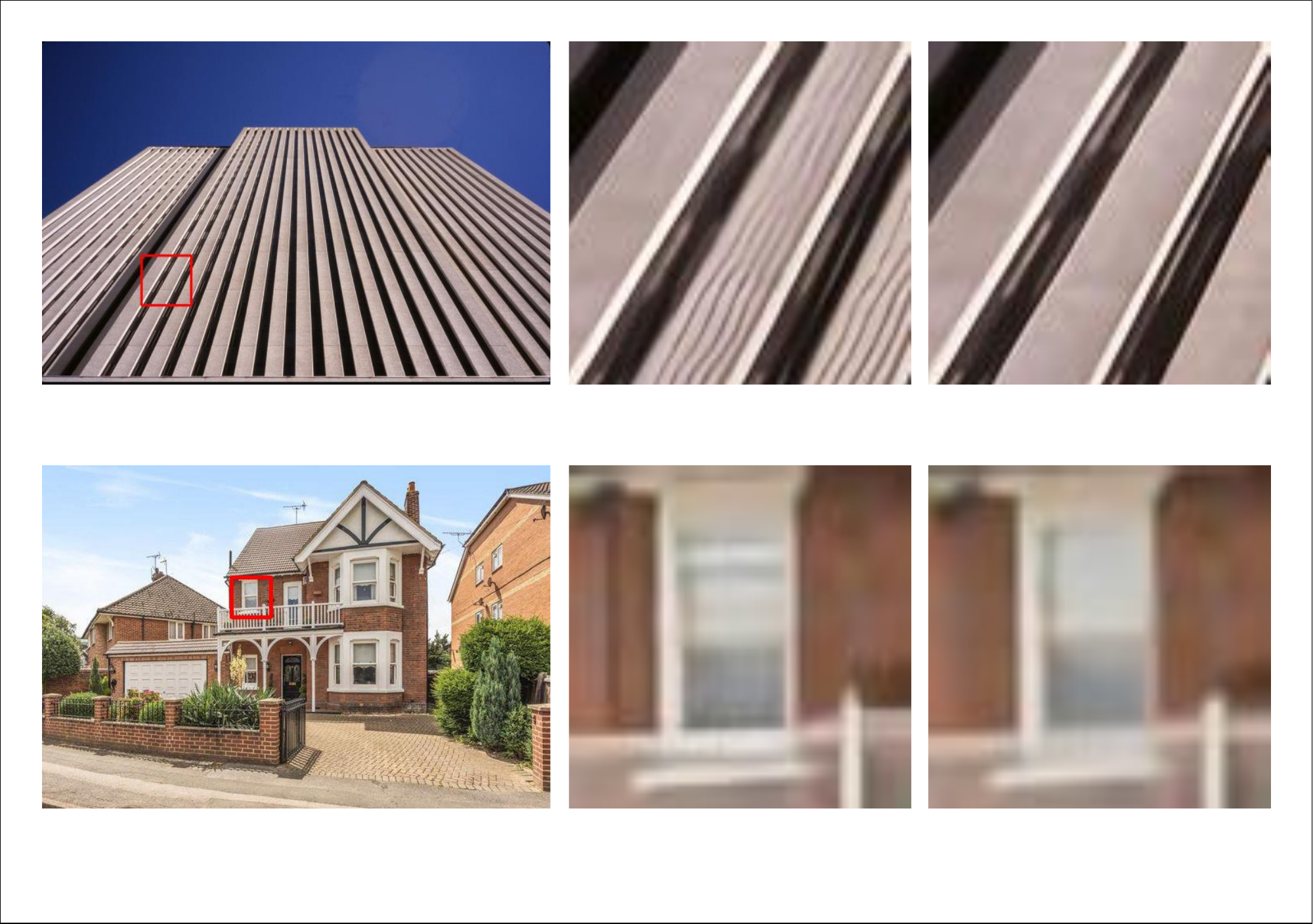}
\put(16,34.5){\color{black}{\small Input}}
\put(16,0.5){\color{black}{\small Input}}
\put(49,34.5){\color{black}{\small \cite{c2}-$rec$}}
\put(49,0.5){\color{black}{\small \cite{datsr}-$rec$}}
\put(78,34.5){\color{black}{\small Ours-$rec$}}
\put(78,0.5){\color{black}{\small Ours-$rec$}}
\end{overpic}
\end{center}%\vspace{-0.1cm}
\caption{Reference-misuse in existing methods.}
\label{fig:wo_irr}
%\vspace{-0.1cm}
\end{figure}
\paragraph{Qualitative Evaluation} Fig.6 shows the visual comparison of our model trained only with reconstruction loss and the existing SISR and RefSR methods. It can be clearly seen that RCAN and RRDB have difficulty in reconstructing texture information due to the severe degradation of high-frequency information, especially for some texts, faces, and some fine textures. Compared with SISR, RefSR can transfer similar textures from the reference images, thus producing more texture details. Compared with some existing RefSR methods, the adaptive nature of FRFSR allows for the perception and transferal of texture information from the Ref images. Thus, the model is capable of compensating for missing high-frequency details in LR, leading to the reconstruction of images with texture details more closely resembling the ground truth. For example, the second pair of local details on the right indicate that RCAN and RRDB fail to reconstruct any blind texture, and the existing RefSR methods generate some texture details, but the images are very unrealistic and far from the ground truth. Our proposed method can generate a sharper, clearer blind texture that is very close to the ground truth. Another example is the fourth pair of images on the left, where the texture is very fine and challenging for RefSR. As can be seen, our method contains more texture details than the existing RefSR methods, which struggle to reconstruct this part and generate very few textures. This demonstrates the effectiveness of our texture search and texture-adaptive aggregation methods. Due to the feature reuse framework, FRFSR can preserve increasingly more realistic texture information when trained with ${{\cal L}^{\textit{rec}}} + {{\cal L}^{\textit{per}}} + {{\cal L}^{\textit{adv}}}$, such as the text on the clothes in the first pair of images on the right in Fig.9, and the stone pillar texture in the fourth pair of images on the left. Compared with the other RefSR methods, our method can generate complete text texture and stone pillar texture, reflecting the advantages of the feature reuse framework and our method.
\begin{table*}[]
\caption{Performance comparison under different relevance levels on CUFED5.\label{tab:table3}}
\centering
\renewcommand\arraystretch{1.5}
\begin{tabular}{c|c|c|c|c|c}
\hline
\multirow{2}{*}{Method} & L1        & L2        & L3        & L4        & LR        \\ \cline{2-6}
                        & PSNR/SSIM & PSNR/SSIM & PSNR/SSIM & PSNR/SSIM & PSNR/SSIM \\ \hline \hline
Cross-Net~\cite{crossnet}         & 25.48/0.764 & 25.48/0.764 & 25.47/0.763 & 25.46/0.763 & 25.46/0.763 \\ \hline
SRNTT-$rec$~\cite{srntt}          & 26.15/0.781 & 26.04/0.776 & 25.98/0.775 & 25.95/0.774 & 25.91/0.776 \\ \hline
SSEN-$rec$~\cite{ssen}            & 26.78/0.791 & 26.52/0.783 & 26.48/0.782 & 26.42/0.781 & - \\   \hline
TTSR-$rec$~\cite{ttsr}            & 26.99/0.800 & 26.74/0.791 & 26.64/0.788 & 26.58/0.787 & 26.43/0.782 \\ \hline
MASA-$rec$~\cite{masa}            & 27.35/0.814 & 26.92/0.796 & 26.82/0.793 & 26.74/0.790 & 26.59/0.784 \\ \hline
${{C}^2}$-Matching-$rec$~\cite{c2}& 28.24/0.841 & 27.39/0.813 & 27.17/0.806 & 26.94/0.799 & 26.53/0.784 \\ \hline
WTRN-$rec$~\cite{wtrn}            & 27.23/0.807 & 26.90/0.794 & 26.79/0.792 & 26.71/0.789 & 26.52/0.783 \\ \hline
TADE-$rec$~\cite{tade}            &28.64/0.850 & 27.77/0.821 &27.46/0.815 &27.23/0.807 &26.83/0.794 \\ \hline
DATSR-$rec$~\cite{datsr}          & 28.50/0.850 & 27.47/0.820 & 27.22/0.811 & 26.96/0.803 & - \\ \hline
Ours-$rec$                        & \bf{29.01/0.860} & \bf{28.01/0.831} & \bf{27.77/0.824} & \bf{27.49/0.815} & \bf{26.88/0.795} \\ \hline
\end{tabular}
\end{table*}

\begin{table}[h]
\caption{Robustness comparison for irrelevant texture transfer on CUFEDR testing set.\label{tab:table4}}
\centering
\renewcommand\arraystretch{1}
\begin{tabular}{|c|c|}
\hline
Method          & \begin{tabular}[c]{@{}c@{}}CUFEDR\\ PSNR/SSIM\end{tabular} \\ \hline
TTSR-$rec$~\cite{ttsr}             & 26.40/0.778                                                \\ \hline
MASA-$rec$~\cite{masa}             & 26.59/0.784                                                \\ \hline
${{C}^2}$-Matching-$rec$~\cite{c2} & 26.50/0.784                                                \\ \hline
DATSR-$rec$~\cite{datsr}           & 26.43/0.784                                                \\ \hline
Ours-$rec$                         & 26.88/0.795                                                \\ \hline
\end{tabular}
\end{table}
\paragraph{Comparison of Robustness of Texture Transformations} Texture transfer robustness is an important criterion for evaluating the performance of RefSR models. As shown in Fig.7, SOTA methods suffer from texture mis-transfer. Moreover, even if the texture of the Ref image is irrelevant, the model should exhibit good adaptive texture transfer robustness. CUFED5 provides four reference images with different levels of relevance (L1-L4). Table 3 shows the results of different models under different relevance settings, where LR denotes the LR image has been used as a reference image. The results demonstrate that our model surpasses several existing RefSR models in terms of texture transfer and robustness. Notably, especially when the reference image is least relevant, our model achieves a performance gain of 0.26db over other SOTA models. Even if the reference image is LR itself, our model
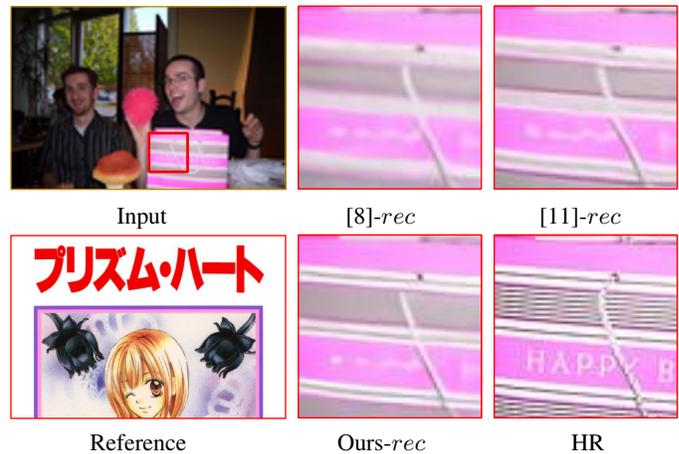
\begin{figure}[!tbp]
\begin{center}
\begin{overpic}[width=0.49\textwidth]{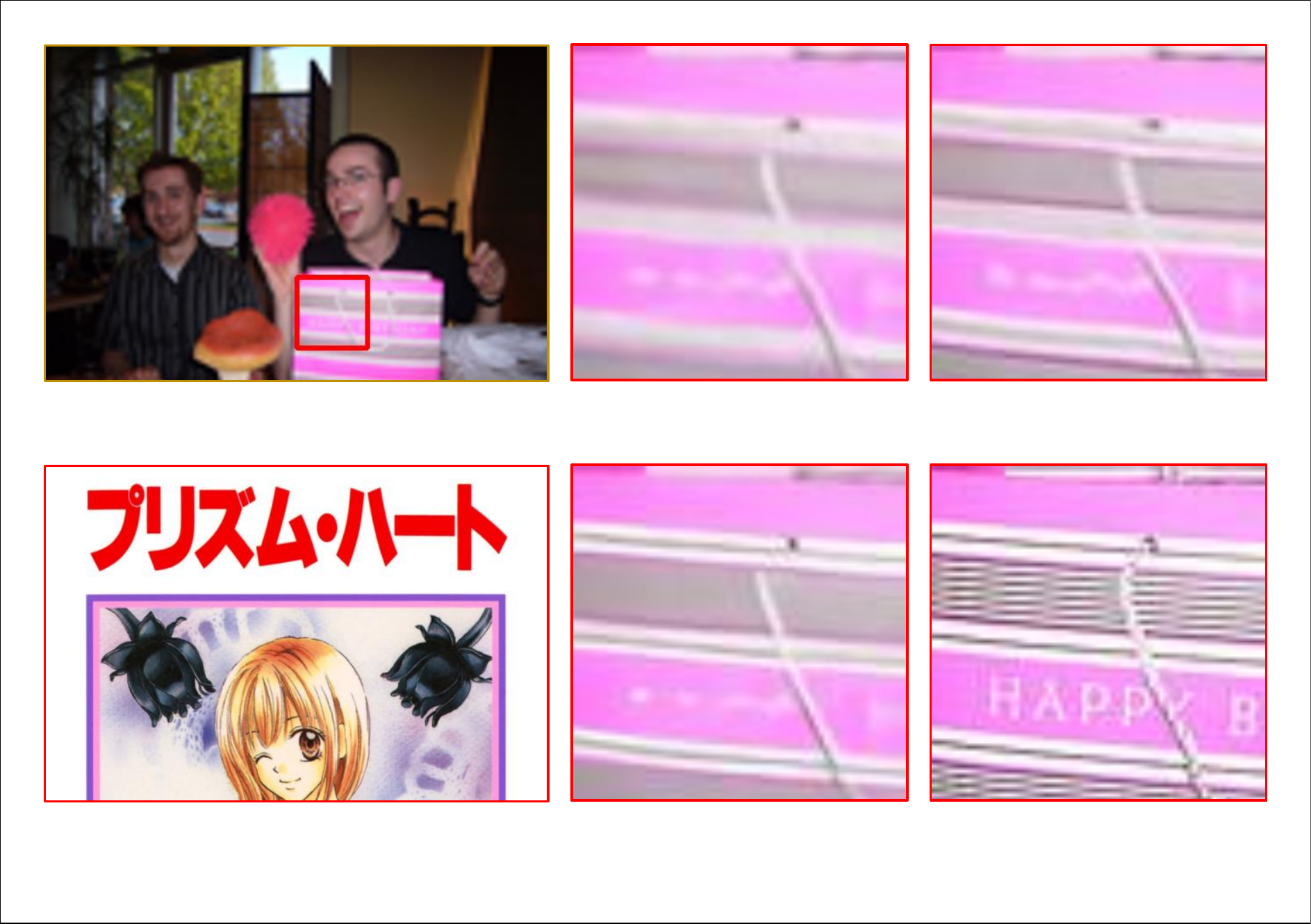}
\put(16,34.5){\color{black}{\small Input}}
\put(12,0.5){\color{black}{\small Reference}}
\put(49,34.5){\color{black}{\small \cite{c2}-$rec$}}
\put(49,0.5){\color{black}{\small Ours-$rec$}}
\put(78,34.5){\color{black}{\small \cite{datsr}-$rec$}}
\put(84,0.5){\color{black}{\small HR}}
\end{overpic}
\end{center}%\vspace{-0.1cm}
\caption{Image reconstruction results under unrelated reference image.}
\label{fig:wo_irr}
%\vspace{-0.1cm}
\end{figure}
can adaptively extract texture information from LR itself and assist super-resolution reconstruction. To further validate the superiority of our model, we created the CUFEDR dataset, which extended all HR images from Urban100, Manga109 and Sun80 into a reference set, consisting of 289 images. During testing, we randomly selected one HR from CUFEDR as a reference image for testing, and the test results are shown in Table 4. Even if the reference image is irrelevant, our model outperforms ${{C}^2}$-Matching by 0.37db, as shown in Fig.8. These experiment results indicate that our model can match and transfer similar textures from relevant reference images, and also possesses adaptive texture transfer robustness in low-relevance scenarios.

\begin{figure*}[t] \centering \subfloat{ \includegraphics[width=0.48\textwidth]{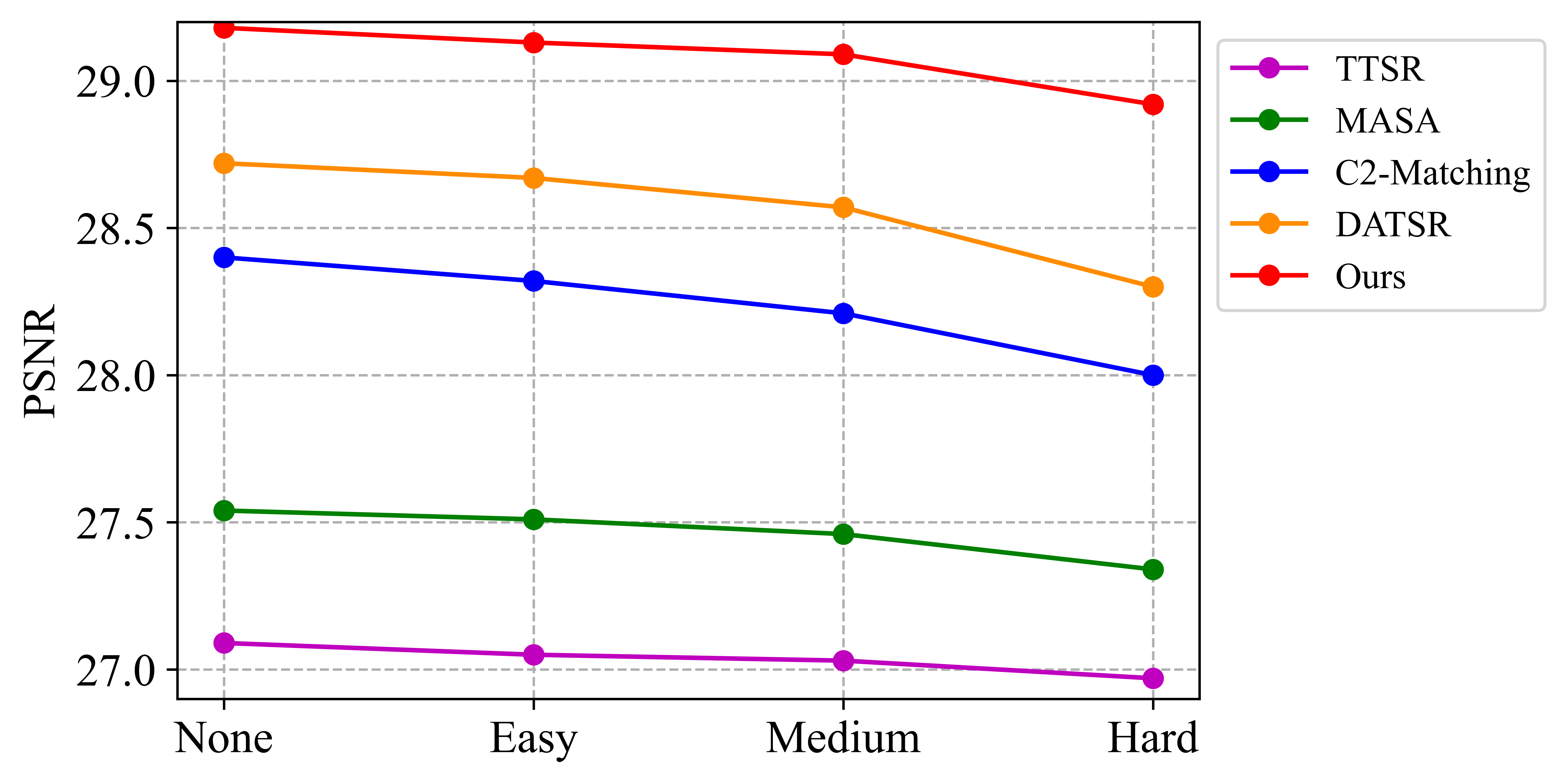} } \subfloat{ \includegraphics[width=0.48\textwidth]{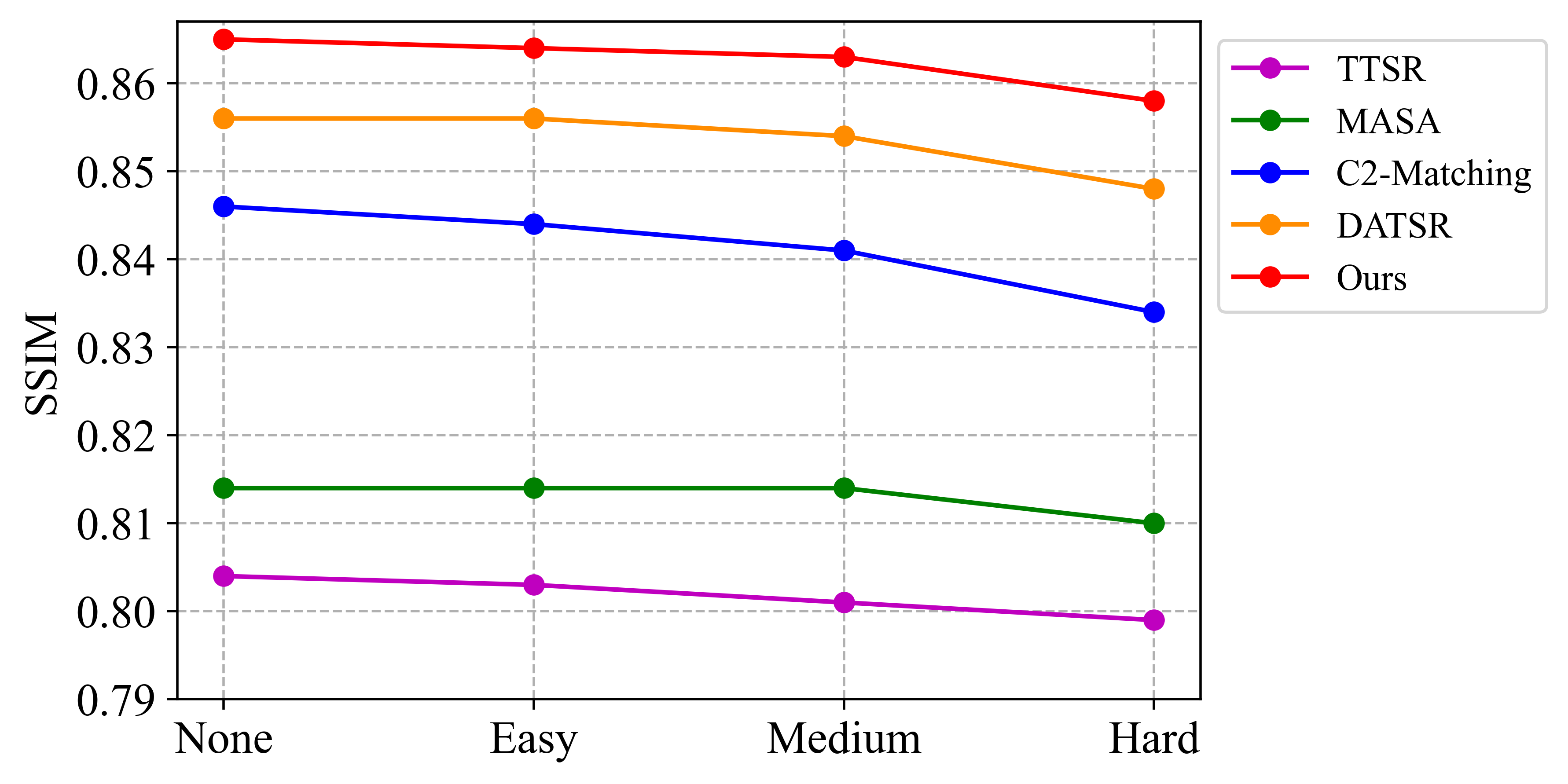} } \caption{Robustness of different models in long distance feature alignment. Our FRFSR is better than TTSR, MASA and ${{C}^2}$-Matching with varying levels of graphic clutter.} \end{figure*}
\begin{figure}[!tbp]
  \centering
  \footnotesize
  \setlength{\tabcolsep}{0pt}
  \begin{tabular}{cccc}
    \includegraphics[width=0.116\textwidth]{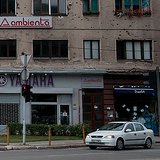}~
    &\includegraphics[width=0.116\textwidth]{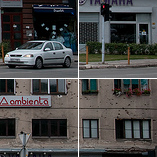}~
    &\includegraphics[width=0.116\textwidth]{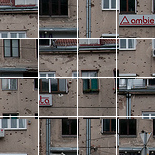}~
    &\includegraphics[width=0.116\textwidth]{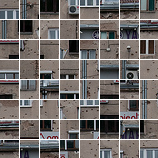}~\\
    None & Easy & Medium & Hard
  \end{tabular}
  \caption{Diagram of different degrees of random disruption.}
  \label{fig:shuffle_patch}
\end{figure}
\paragraph{Comparison of Robustness of Long-range Alignment} To enhance the robustness of our model with respect to long-distance feature alignment, we integrated training with long-distance alignment and context perturbation samples. Specifically, we divide the reference image into multiple n$\times$n patches, then randomly shuffle their positions, and finally reassemble them into a new complete sample image, which disturbs the contextual dependency of the image and enlarges the misalignment distance between relevant patches. During testing, we perform three different levels of random shuffling on the reference image, namely easy, medium, and hard, which divide the image into 2$\times$2, 4$\times$4, and 8$\times$8 patches respectively, as shown in Fig.10. Fig.11 shows our model and the ${{C}^2}$-Matching method for different levels of randomly shuffled reference images. By using cross-layer semantic regularization to fuse and enhance texture features with similar semantics at different granularities, we show that our model is more robust than the ${{C}^2}$-Matching method. It is worth noting that we only use medium-level data augmentation during training.
\subsection{Discussion of Model Size and Computation Cost}
In this section, we compare our method with other methods in terms of parameter size and running time, as shown in Table 5 and Table 6. Our model improves the running time by 33.5\% and reduces the parameters by 25\% compared to DATSR~\cite{datsr}, which is based on SwinTransformer~\cite{swint}as the basic module. However, compared to MASA's~\cite{masa}coarse-to-fine matching method, our method increases both running time and parameters, but our method greatly improves the performance. We selected some models for a visual comparison of parameters, running time and model performance, as shown in Fig.12. Note that our method is the model performance and parameter size after removing the SIFE module. Since our model consists of two parts: relevant texture search and transfer with SIFE, the model parameters are relatively large. The SIFE module can reduce the introduction of some texture error, so it plays an auxiliary role. It is worth noting that after removing the SIFE module, the parameters are only 13.5M, but the performance can still reach SOTA.
\begin{table}[h]
    \caption{Running time of our FRFSR compared with several other RefSR models on CUFED5}\label{tab:my_labe5}
    \centering
    \begin{tabular}{p{5.1cm}|lc}
    \hline
        \multicolumn{1}{l|}{Model} & \multicolumn{1}{|c}{Runtime(ms)} \\ \hline
        SRNTT~\cite{srntt}                       & 13256                            \\
        TTSR~\cite{ttsr}                         & 505                              \\
        MASA~\cite{masa}                         & 336                              \\
        $C^2$-Matching~\cite{c2}                 & 361                              \\
        DATSR~\cite{datsr}                       & 1214                             \\
        Ours                                     & 807                             \\ \hline
    \end{tabular}
\end{table}

\begin{table}[h]
    \caption{Comparison between our FRFSR and several other RefSR models in terms of the number of parameters.}\label{tab:my_labe6}
    \centering
    \begin{tabular}{l|c|c}
            \hline
        \multicolumn{1}{l|}{Model} & Params      & \multicolumn{1}{|c}{PSNR↑/SSIM↑} \\ \hline
        CorssNet~\cite{crossnet}                    & 33.18M      & 25.48/0.764                    \\
        TTSR-$rec$~\cite{ttsr}                      & 6.4M        & 27.09/0.804                    \\
        MASA-$rec$~\cite{masa}                      & 4.0M        & 27.54/0.814                    \\
       ${{C}^2}$-Matching-$rec$~\cite{c2}           & 8.9M        & 28.24/0.841                    \\
        TADE-$rec$ (w/o Decouple)~\cite{tade}        & 10.9M       & 28.36/0.842                    \\
        TADE-$rec$~\cite{tade}                      & 10.9M+15.9M & 28.64/0.850                    \\
        DATSR-$rec$~\cite{datsr}                    & 18M         & 28.72/0.856                    \\
        Ours-$rec$ (w/o SIFE)                        & 13.5M       & 28.93/0.865                    \\
        Ours-$rec$                                  & 13.5M+15.9M & 29.16/0.865                    \\ \hline
    \end{tabular}
\end{table}

\begin{figure}[!t]
\centering
\includegraphics[width=0.49\textwidth]{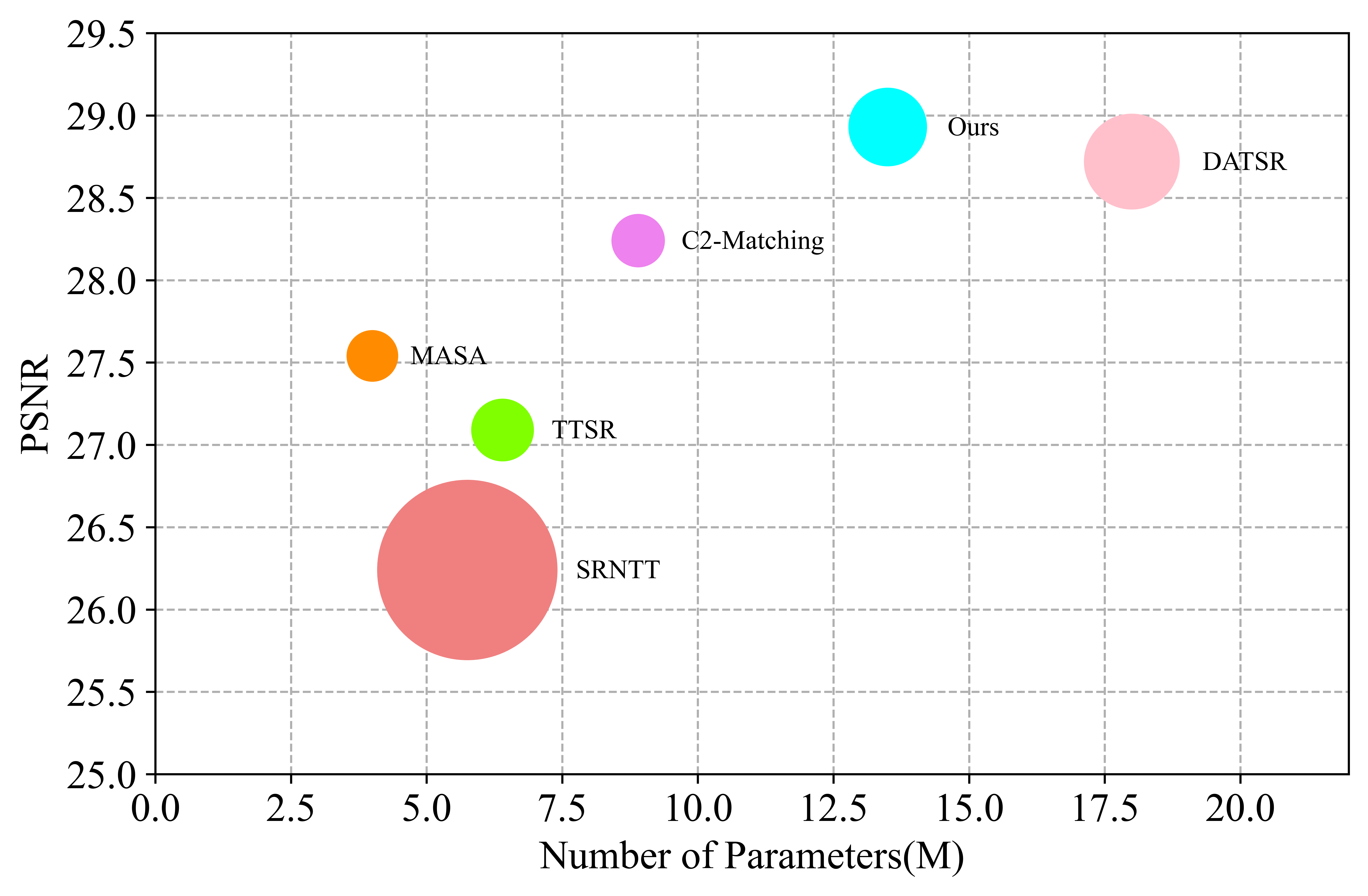}
\caption{Visual comparison of parameters, running time and performance of some models. The size of the circle indicates the running time. It is worth noting that our method in the figure is without the SIFE module for performance and parameter size.}
\label{fig_1}
\end{figure}
\subsection{Ablation Studies}
In this section, we evaluate our method’s dynamic residual block component and single image feature embedding using CUFED5. Table 7 shows the evaluation results. We also apply the feature reuse framework to other RefSR methods to demonstrate its effectiveness.
\begin{table}[h]
\caption{Quantitative evaluation of the ablation study on the single-image feature embedding module and dynamic residual ResBlock component on the CUFED5.}\label{tab:my_labe7}
\centering
\begin{tabular}{l|c|c|c|c}
\hline
\multicolumn{1}{l|}{Model} & SIFE & DRB & PSNR↑/SSIM↑ & LPIPS↓ \\ \hline
Baseline                    &      &              & 28.68/0.853 & 0.1630 \\
Baseline+DRB                &     &\checkmark     & 28.93/0.865 & 0.1563 \\
Baseline+SIFE               &\checkmark       &   & 29.01/0.861 & 0.1462 \\
Baseline+DRB+SIFE           & \checkmark    & \checkmark   & 29.16/0.865 & 0.1431 \\ \hline
\end{tabular}
\end{table}

\begin{figure}[!tbp]
\begin{center}
\begin{overpic}[width=0.49\textwidth]{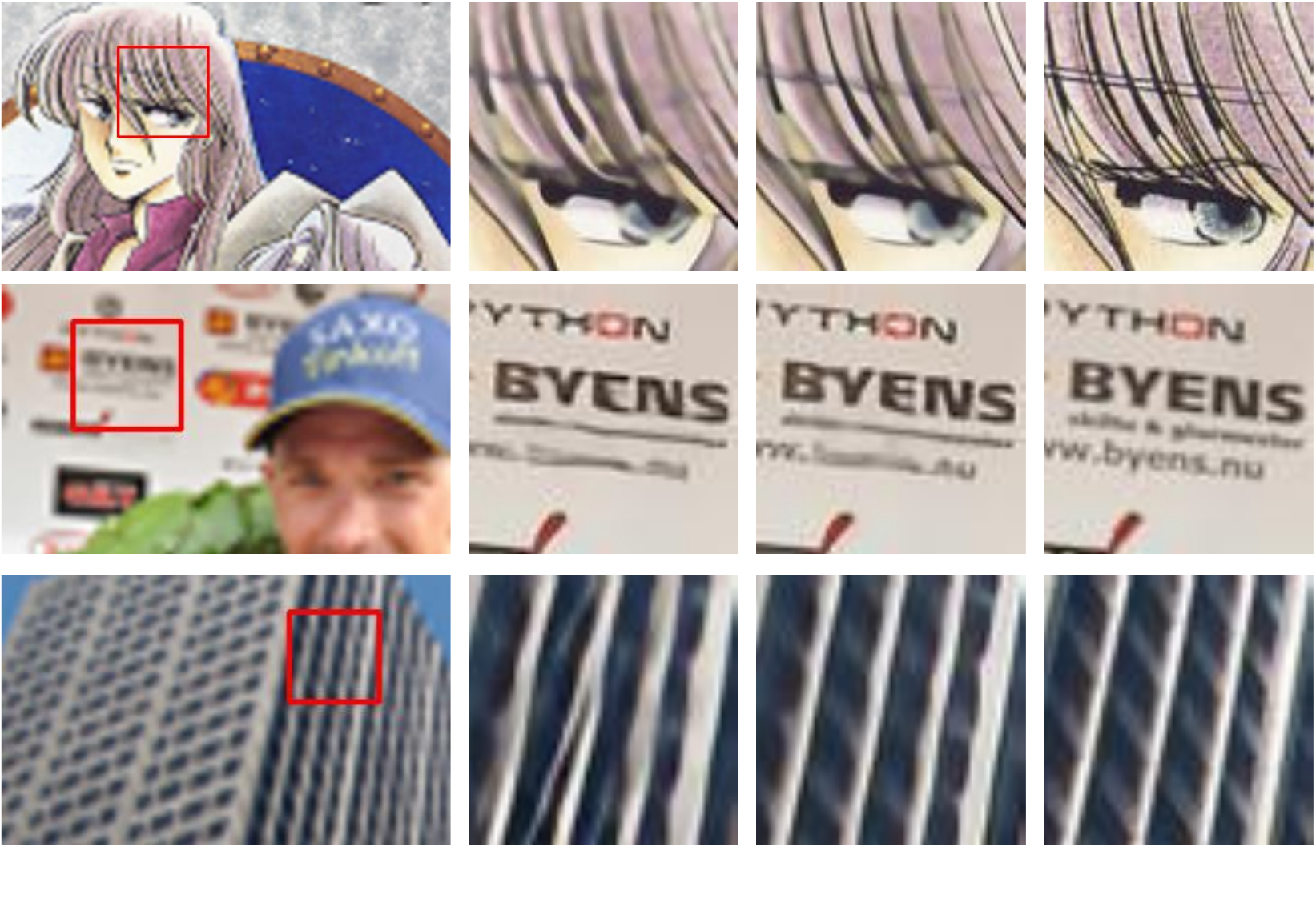}
\put(13.5,1){\color{black}{\small Input}}
\put(38.5,1){\color{black}{\small w/o SIFE}}
\put(61,1){\color{black}{\small w/ SIFE}}
\put(87,1){\color{black}{\small HR}}
\end{overpic}
\end{center}
\caption{Ablation analysis of the SIFE module is carried out on Urban100, CEUFED5, and Sun80. In addition to boosting the performance of texture transfer and reconstruction, the SIFE module effectively suppresses irrelevant textures from being introduced.}
\label{fig:sife_ab}
%\vspace{-0.1cm}
\end{figure}
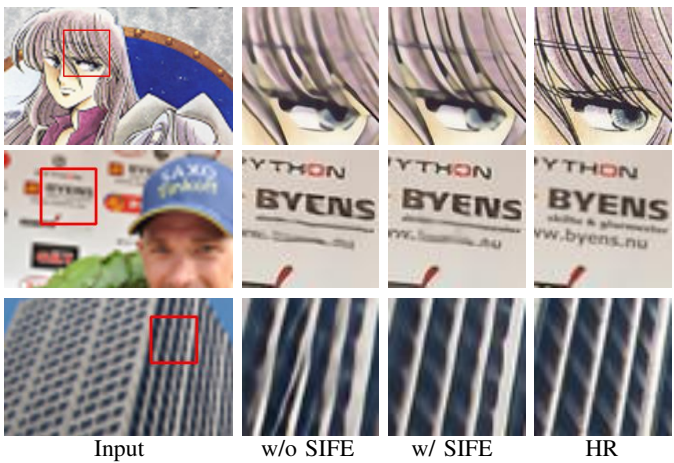

\begin{table}[h]
\caption{Quantitative ablation experiments of SIFE on multiple benchmark datasets. In addition, we also chose the decoupled and non-decoupled frameworks in TADE for comparison.}\label{tab:my_labe7}
\begin{tabular}{c|c|c|c|c}
\hline
Method                                                    & CUFED5      & Sun80       & Urban100    & Manga109   \\ \hline
\begin{tabular}[c]{@{}c@{}}TADE~\cite{tade}\\(Couple)\end{tabular} & 28.36/0.842 & 30.01/0.812 & 25.66/0.769 & 29.98/0.901 \\ \hline
\begin{tabular}[c]{@{}c@{}}Ours\\ (w/o SIFE)\end{tabular}                                                  & 28.68/0.853 & 30.07/0.814 & 25.95/0.779 & 30.15/0.907 \\ \hline
\begin{tabular}[c]{@{}c@{}}TADE~\cite{tade}\\ (Decouple)\end{tabular}                                                   & 28.64/0.850 & 30.31/0.820 & 26.71/0.807 & \bf{31.23/0.917} \\ \hline
\begin{tabular}[c]{@{}c@{}}Ours\\ (w/ SIFE)\end{tabular}                                                   & \bf{29.01/0.861} & \bf{30.35/0.821} & \bf{26.80/0.810} & 31.14/0.916 \\ \hline
\end{tabular}
\end{table}
\paragraph{Single Image Feature Embedding} The reconstructed features from SISR can effectively compensate for the remaining features other than the texture features in the Ref image. To verify the effectiveness of single-image feature embedding, we do not consider feature embedding when transferring texture, and we find that both the performance of transferring matching texture from the Ref image and reconstructing a similar texture that does not exist in the Ref image are affected. Table 8 shows that with the help of the SIFE module, our model not only improved by 0.37db on CUFED5, but also achieved corresponding improvements on other datasets. In addition, we also compared our model with the decoupled and coupled frameworks in TADE~\cite{tade} at the same time, further demonstrating the effectiveness of SIFE. On the other hand, as shown in Fig.13, adding the SIFE module can facilitate the model's learning, alleviate detail loss, and not only transfer richer and finer texture details from the Ref images in CUFED5, it can also make texture features more prominent in the SR images reconstructed on other datasets, such as the hair texture of anime images. It is worth noting that adding the SIFE module can suppress irrelevant texture transfer to some extent, as shown in the third row. Through quantitative and qualitative evaluation, the SIFE module improves the model's ability to transfer texture and recover texture details that do not exist in the Ref image.
\begin{figure}[!tbp]
\begin{center}
\begin{overpic}[width=0.49\textwidth]{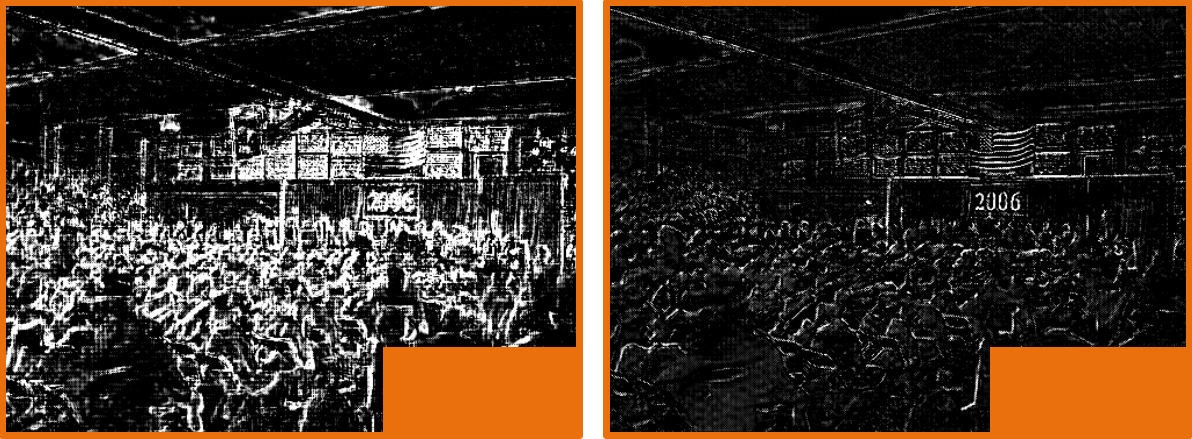}
\put(34,3.2){\color{white}{\small w/o ESA}}
\put(86,3.2){\color{white}{\small w/ ESA}}
\end{overpic}
\end{center}%\vspace{-0.1cm}
\caption{Feature visualization of Enhanced Spatial Attention (ESA). It can effectively reduce noisy textures while enhancing similar textures.}
\label{fig:esa_ab}
%\vspace{-0.1cm}
\end{figure}

\begin{figure*}[htbp!]
\centering
\footnotesize
\renewcommand\arraystretch{1.2}
\begin{tabular}{p{0.21\textwidth}<{\centering} p{0.075\textwidth}<{\centering} p{0.04\textwidth}<{\centering} p{0.085\textwidth}<{\centering} p{0.22\textwidth}<{\centering} p{0.06\textwidth}<{\centering} p{0.06\textwidth}<{\centering} p{0.06\textwidth}<{\centering}}
LR        & RCAN & TTSR  & MASA & LR & ESRGAN & SRNTT & MASA \\ \cline{1-8}
Reference & ~\cite{c2}   & DATSR & Ours   & Reference & ~\cite{c2}   & DATSR & Ours
\end{tabular}
\label{table:example}

\includegraphics[width=\textwidth]{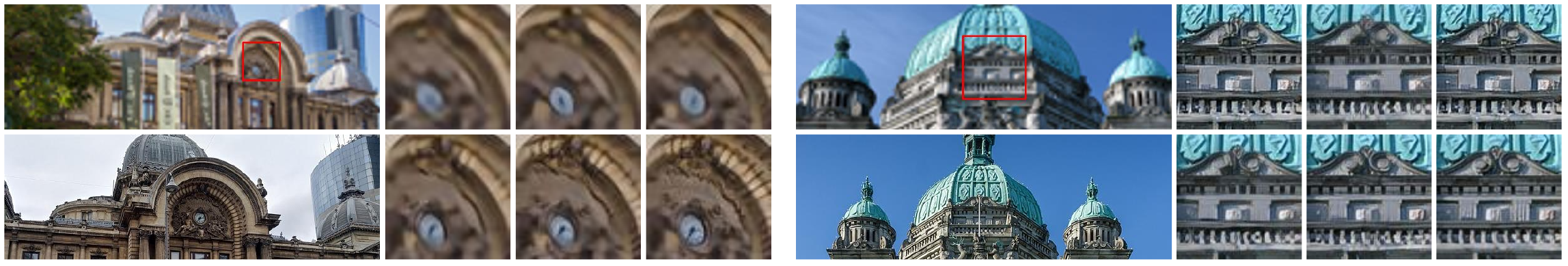}
\caption{Web search applications on multiple methods. The pair of images presented on the left show the SR images obtained by only training with the reconstruction loss (-$rec$), while the ones on the right were obtained by training with all losses.}
\label{fig:results}
\end{figure*}

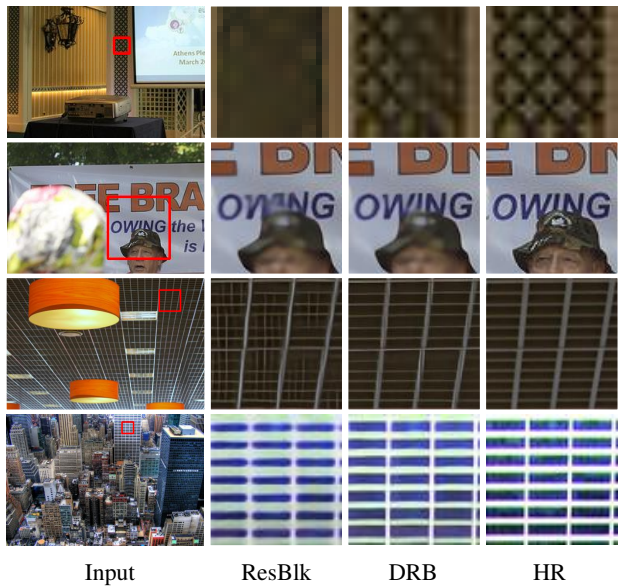
\begin{figure}[!htbp]
\begin{center}
\begin{overpic}[width=0.45\textwidth]{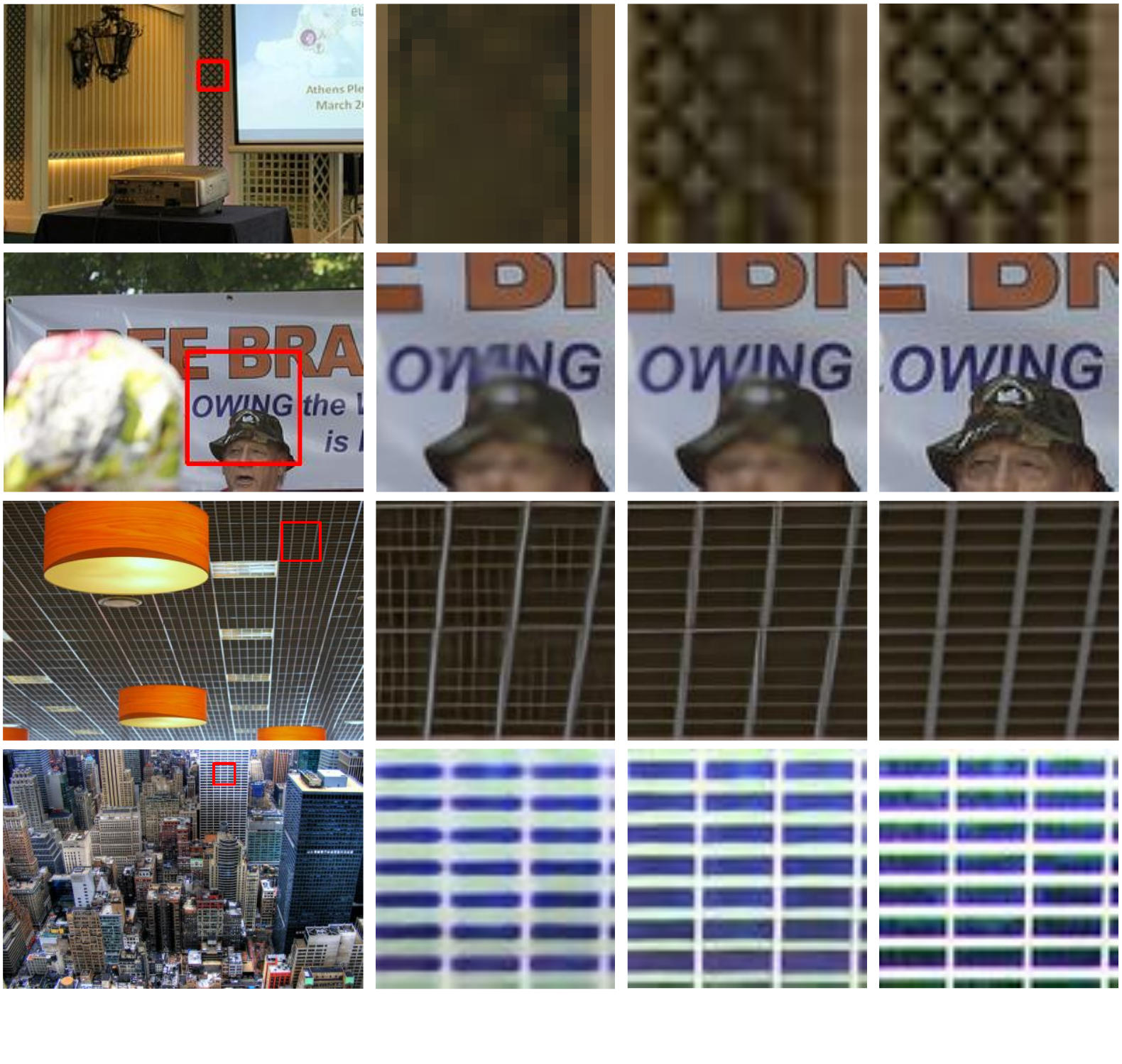}
\put(13,1){\color{black}{\small Input}}
\put(38.5,1){\color{black}{\small ResBlk}}
\put(62.5,1){\color{black}{\small DRB}}
\put(86,1){\color{black}{\small HR}}

\end{overpic}
\end{center}%\vspace{-0.1cm}
\caption{Qualitative ablation study on the DRB module.}
\label{fig:drb_ab}
%\vspace{-0.1cm}
\end{figure}
\paragraph{Dynamic Residual Block}
The aligned reference features contain a lot of noise information, and using ResBlock to directly aggregate the reference features will cause the SR image to have irrelevant textures and noise. We add dynamic filters and enhanced spatial attention (ESA) in the residual block, which can effectively perceive relevant textures and adaptively aggregate them. On CUFED5, compared with Baseline, the model with DRB has a 0.25db improvement on PSNR, as shown in Table 7. Fig.14 shows the feature visualization of ESA. It can be seen that after adding ESA, the texture features with higher relevance become more prominent, and the texture edges become sharper. Furthermore, after adding DRB, the LPIPS~\cite{lpips} also decreased by 0.0067, and the smaller LPIPS corresponds to better performance. The third rows of Fig.16 show that adding the DRB module can effectively reduce the introduction of irrelevant textures and enhance the model's ability to transfer texture and reconstruct texture. The fourth row of Fig.16 shows that the texture reconstructed by the DRB module is closer to the ground truth.
\begin{table}[]
\renewcommand\arraystretch{1.5}
\caption{Quantitative ablation study on adding FRF on multiple methods.}\label{tab:my_labe7}
\centering
\begin{tabular}{l|c|c|c}
\hline
\multicolumn{1}{c|}{Model} & FRF & PSNR↑/SSIM↑ & LPIPS↓ \\ \hline
MASA~\cite{masa}                        &     & 24.92/0.729 & 0.0987 \\
MASA+FRF                    & \checkmark    & 25.16/0.744 & 0.0954 \\
${{C}^2}$-Matching~\cite{c2}                 &    & 27.16/0.805 & 0.1229 \\
${{C}^2}$-Matching+FRF             & \checkmark    & 28.05/0.834 & 0.1198 \\
Ours                        &     & 28.29/0.840 & 0.0992 \\
Ours+FRF                    & \checkmark    & 28.71/0.852 & 0.0974 \\ \hline
\end{tabular}
\end{table}
\paragraph{Feature Reuse Framework} We found that compared with the model trained with reconstruction loss, the model trained with all loss exhibited a worse performance on texture transfer and reconstruction. To reduce the impact of adversarial loss and perceptual loss, we used a Feature Reuse Framework (FRF) to supplement the texture that could not be reconstructed. Table 9 shows the effect of FRF on MASA and $C^2$-Matching. It can be seen that after adding FRF, all models consistently improved. Although PSNR and SSIM cannot determine visual quality, we also use LPIPS as an evaluation indicator. It is noted that our model is slightly lower than MASA on LPIPS, which is because MASA uses larger weights for adversarial loss and perceptual loss, resulting in better visual effects for the final output. But after adding FRF, PSNR and SSIM still increased by 0.24db and 0.015 respectively, and LPIPS decreased by 0.0033, indicating the effectiveness of this training strategy. Our loss weights are consistent with $C^2$-matching, but compared with $C^2$-Matching with FRF added, our model improved by 0.66db and 0.018 on PSNR and SSIM respectively, and LPIPS decreased by 0.0224. We visualize the methods with FRF added, and the visualization results are shown in Fig.17. It is noted that the model trained with only reconstruction loss (-$rec$) has more texture details than the model trained with all losses, such as the wire mesh in MASA and the American flag in $C^2$-matching. However, after adding FRF, the texture can be restored to normal. This indicates that this framework can reduce the impact of adversarial loss and perceptual loss on texture reconstruction degradation.
\begin{figure}[!htbp]
\footnotesize
	\centering
\setlength{\tabcolsep}{1.4pt}
\begin{tabular}{ccccccc}
    \multicolumn{2}{c}{
        \includegraphics[width=0.10\textwidth]{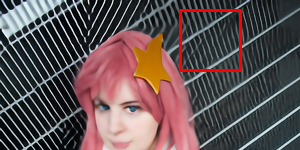}
		\includegraphics[width=0.05\textwidth]{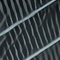}} &
    \multicolumn{2}{c}{
		\includegraphics[width=0.10\textwidth]{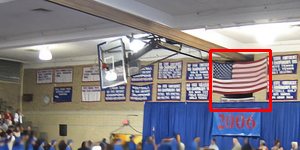}
        \includegraphics[width=0.05\textwidth]{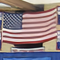}}   &
    \multicolumn{2}{c}{
        \includegraphics[width=0.10\textwidth]{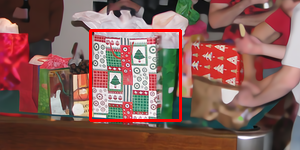}
		\includegraphics[width=0.05\textwidth]{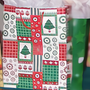}}\\

  \multicolumn{2}{c}{MASA-$rec$~\cite{masa}} & \multicolumn{2}{c}{$C^2$-Matching-$rec$~\cite{c2}} & \multicolumn{2}{c}{\textbf{FRFSR}-$rec$(Ours)} \\

    \multicolumn{2}{c}{
        \includegraphics[width=0.10\textwidth]{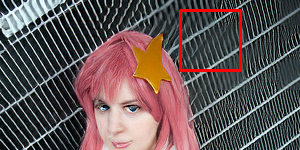}
		\includegraphics[width=0.05\textwidth]{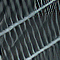}} &
    \multicolumn{2}{c}{
		\includegraphics[width=0.10\textwidth]{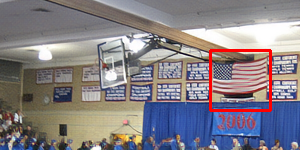}
        \includegraphics[width=0.05\textwidth]{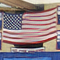}}   &
    \multicolumn{2}{c}{
        \includegraphics[width=0.10\textwidth]{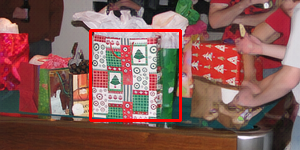}
		\includegraphics[width=0.05\textwidth]{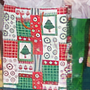}}\\

  \multicolumn{2}{c}{MASA} & \multicolumn{2}{c}{$C^2$-Matching} & \multicolumn{2}{c}{\textbf{FRFSR}(Ours)} \\

    \multicolumn{2}{c}{
        \includegraphics[width=0.10\textwidth]{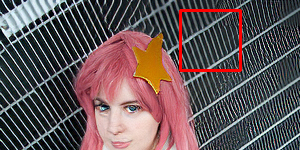}
		\includegraphics[width=0.05\textwidth]{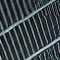}} &
    \multicolumn{2}{c}{
		\includegraphics[width=0.10\textwidth]{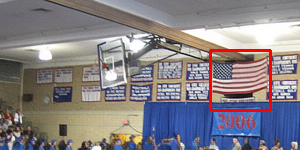}
        \includegraphics[width=0.05\textwidth]{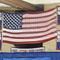}}   &
    \multicolumn{2}{c}{
        \includegraphics[width=0.10\textwidth]{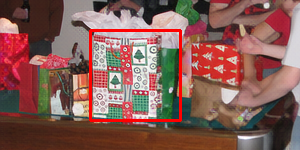}
		\includegraphics[width=0.05\textwidth]{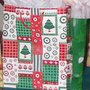}}\\

  \multicolumn{2}{c}{MASA+FRF} & \multicolumn{2}{c}{$C^2$-Matching+FRF} & \multicolumn{2}{c}{\textbf{FRFSR}(Ours)+FRF} \\

	\end{tabular}
	\caption{Qualitative ablation study on adding FRF (Feature Reuse Framework) to multiple methods, where the first row is the SR results of each model trained with only reconstruction loss, the second row is the output results of models trained with all losses, and the third row is the output results of each model after adding FRF.
}
	\label{fig:visualresults}\vspace{-0.1cm}
\end{figure}

\subsection{Web Search Applications} Image search by image is a common feature of the existing network, which is also one of the most typical applications of RefSR. By searching for reference images based on the user input LR image, RefSR can reconstruct LR. At the same time, this is also a way to verify the generalization ability of RefSR. We selected two low-resolution images from DIV2K and utilized Google's image search by image function to locate corresponding reference images. Our FRFSR method was then compared with other SISR and RefSR methodologies, and the results are presented in Fig.15. Compared with general SISR methods (RCAN and ESRGAN), and existing RefSR methods (SRNTT, TTSR, MASA, $C^2$-Matching, DATSR), our method can transfer more details and textures from the images found in the web, even if there are differences in lighting, texture size or perspective in the reference image. Therefore, the SR image reconstructed by our method has a better visual quality.
\section{Conclusion}
In this paper, we introduce a feature reuse framework that successfully mitigates the negative impacts of the perceptual and adversarial losses that arise during the texture reconstruction process. Our method is composed of two modules: a single-image feature embedding module for reconstructing the self-features of the LR input image, and a texture adaptive aggregation module for reconstructing the effective texture of the perceptual aggregate reference image. Our approach improves robustness to unrelated references. The experiments conducted on various benchmarks show that our approach outperforms existing RefSR methods in both qualitative and quantitative measures.

\bibliographystyle{IEEEtran}
\bibliography{refs}

\begin{IEEEbiography}[{\includegraphics[width=1in,height=1.25in,clip,keepaspectratio]{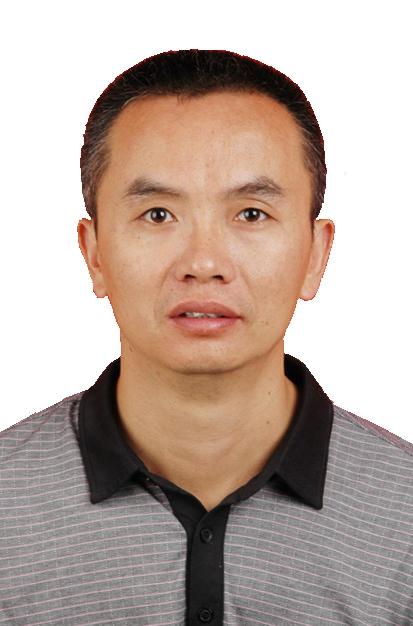}}]{Xiaoyong Mei}received his Ph.D. degrees in computer software and theory from Sun Yat-sen University, Guangzhou, China, in 2010. He is currently a Full Professor with Zhejiang Normal University. He has been working on C image processing and analysis, video analysis and dynamic scene processing, information extraction, and visual applications and systems. His research interests include in explainable machine learning, natural language processing, and computer vision. He has authored or coauthored more than 20 scientific papers in international journals and conferences.
\end{IEEEbiography}

\begin{IEEEbiography}[{\includegraphics[width=1in,height=1.25in,clip,keepaspectratio]{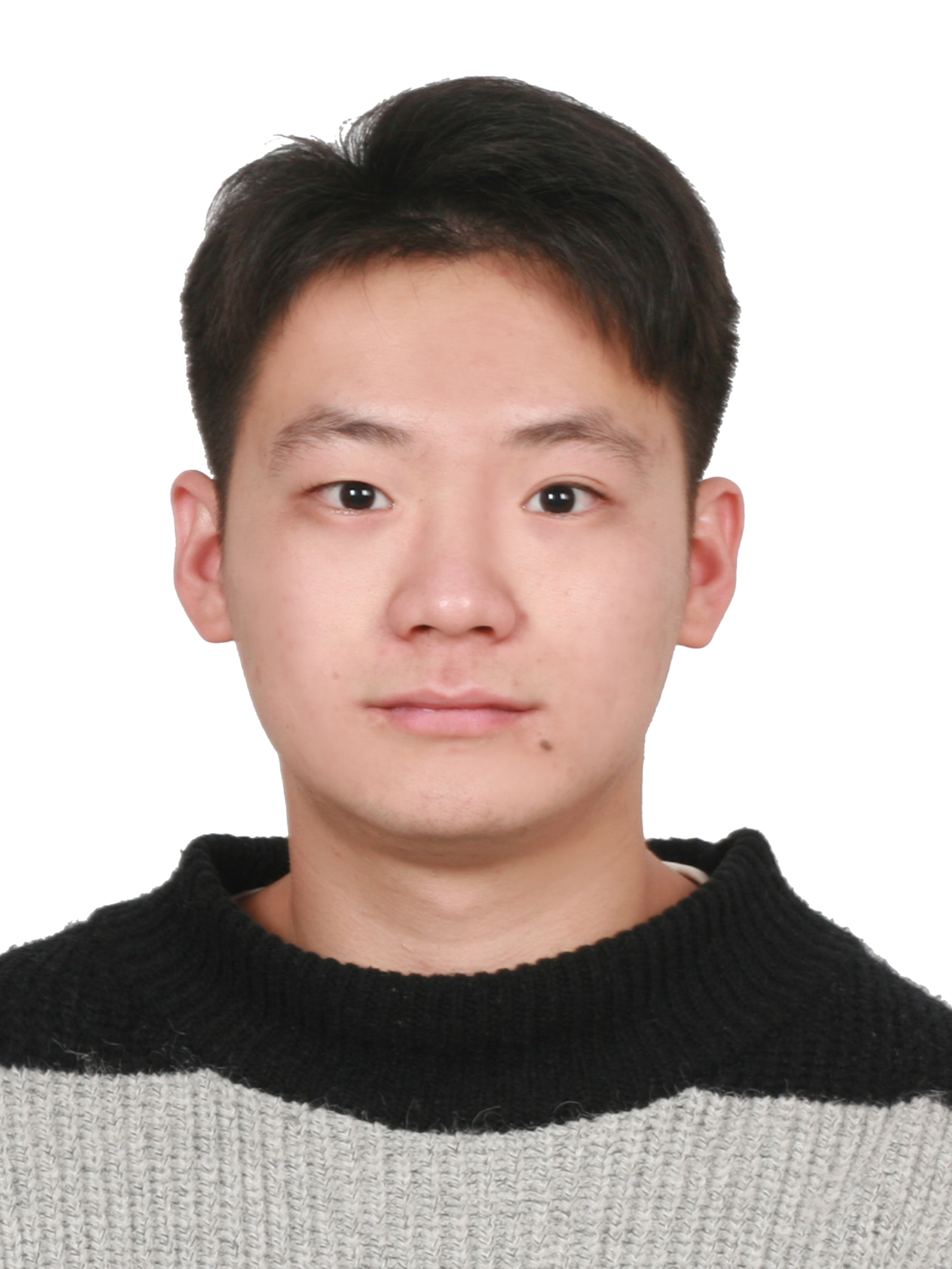}}]{Yi Yang}
received the bachelor's degree in Electronic and Information Engineering from Ningbo University of Technology, Ningbo, Zhejiang, China, in 2021. He is currently pursuing the Master's degreein educational technology with Zhejiang Normal University, Jinhua, China. His research interests include computer vision, image processing, machine learning, etc..
\end{IEEEbiography}

\begin{IEEEbiography}[{\includegraphics[width=1in,height=1.35in,clip,keepaspectratio]{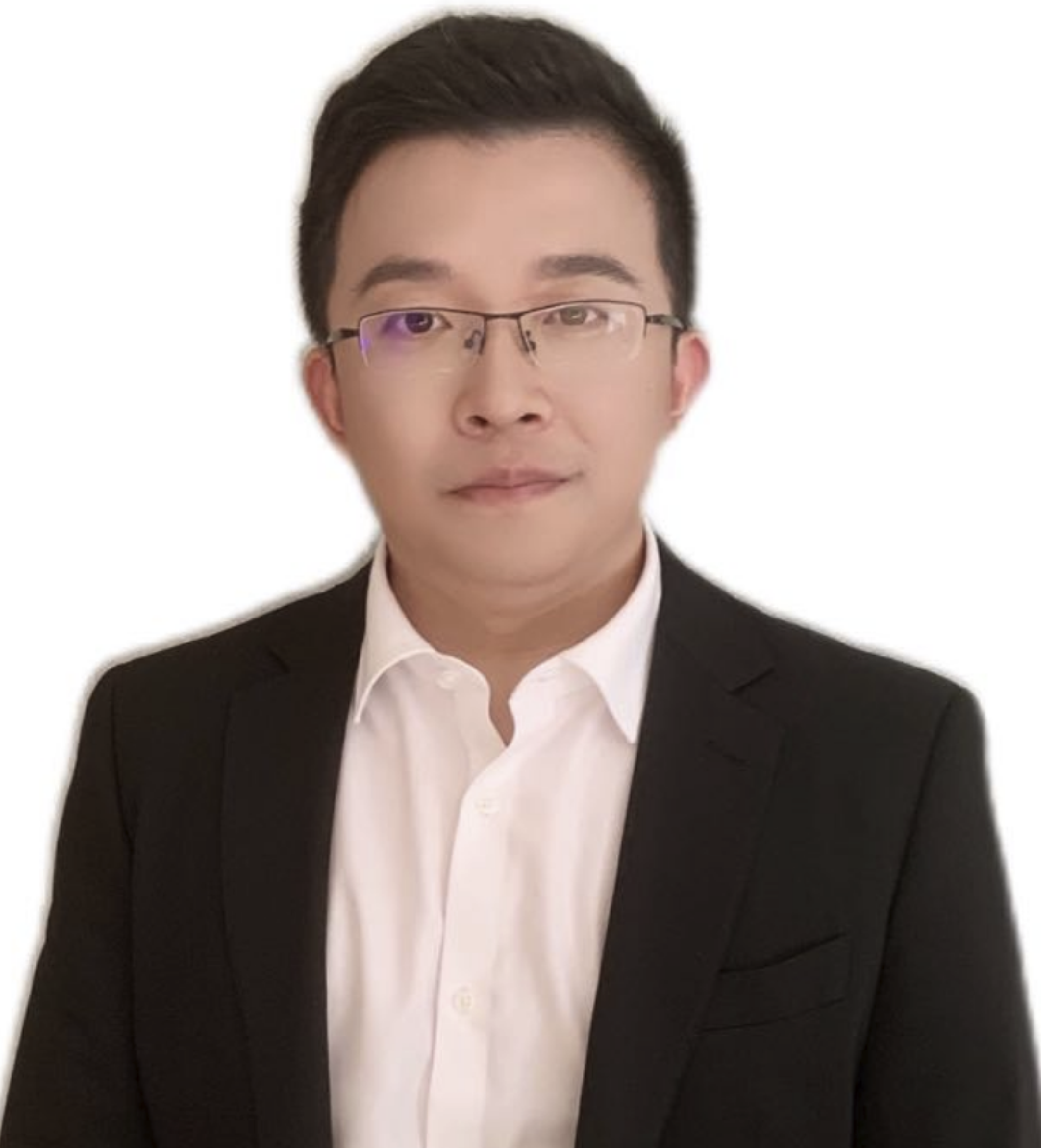}}]{Ming Li}
is currently a ``Shuang Long Scholar'' Distinguished Professor at the Key Laboratory of Intelligent Education Technology and
Application of Zhejiang Province, Zhejiang Normal University, China. He received his PhD degree from the Department of Computer Science
and IT at La Trobe University, Australia, in 2017.
He completed two Postdoctoral Fellowship positions with the Department of Mathematics and
Statistics, La Trobe University, Australia, and the
Department of Information Technology in Education, South China Normal University, China, respectively. He has
published in top-tier journals and conferences, including IEEE TPAMI,
Artificial Intelligence, npj Precison Oncology, IEEE TKDE, IEEE TNNLS, IEEE TCYB,
IEEE TII, IEEE TITS, ACM TMOS, NeurIPS, ICML, IJCAI, etc. He is a regular
reviewer for top journals including IEEE TPAMI, Artificial Intelligence,
IEEE TNNLS, IEEE TCYB (Outstanding Reviewer in 2016 and 2017),
IEEE TKDE, IEEE TII, IEEE TITS, NN, INS, and others. Dr. Li, as a leading guest editor, organized a special issue ``\emph{Deep
Neural Networks for Graphs: Theory, Models, Algorithms and Applications}'' in \emph{IEEE Transactions on Neural Networks and Learning Systems}, and a special session on ``\emph{Recent Advances in
Deep Learning for Graphs}'' in LOD2022. He is a PC member at ICML,
NeurIPS, ICLR, IJCAI, AAAI, KDD, ICLR, an Associate Editor of \emph{Neural
Networks},  an Associate Editor of \emph{Applied Intelligence},  an Associate Editor of \emph{Soft Computing},  and an Associate Editor of \emph{Neural Processing Letters}.
\end{IEEEbiography}

\begin{IEEEbiography}[{\includegraphics[width=1.25in,height=1.583in,clip,keepaspectratio]{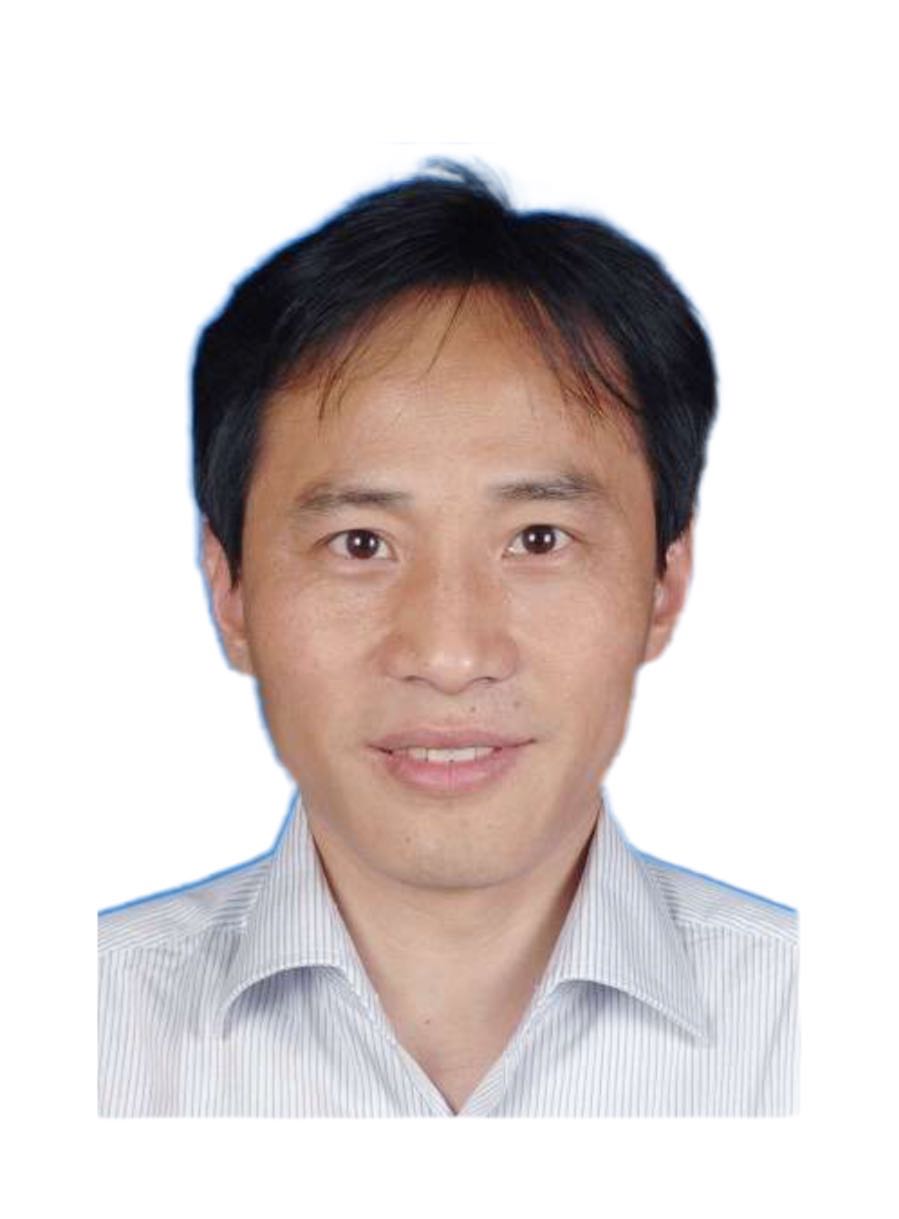}}]{Changqin Huang}
 received the Ph.D. degree in Computer Science and Technology from Zhejiang University, China, in 2005. He completed two Postdoctoral Fellowship positions with the ECNU-TCL Joint Workstation on Educational Technology and the Sun Yat-sen University on Computer Software and Theory. Dr. Huang had completed visiting research at University of California Irvine, USA, in 2011, and La Trobe University, Australia, in 2018. He is currently a Distinguished Professor with Zhejiang Normal University, China, and also the director of the Key Laboratory of Intelligent Education Technology and Application of Zhejiang Province, China. His research interests range from Big Data in Education to Machine Learning and Intelligent Education. He has published several papers in prestigious journals in Computer Science and Educational Technology, including IEEE TPAMI, CAE, IEEE TIP, IEEE TCYB, IEEE TNNLS, IEEE TII, CHB, BJET, etc., and acted as PI for many projects related to AI and Its Applications in Education.

Dr. Huang is a Guangdong Specially-Appointed Professor (Pearl River Scholar). He serves as an Associate Editor of IEEE Transactions on Learning Technologies.

\end{IEEEbiography}

\begin{IEEEbiography}[{\includegraphics[width=1in,height=1.25in,clip,keepaspectratio]{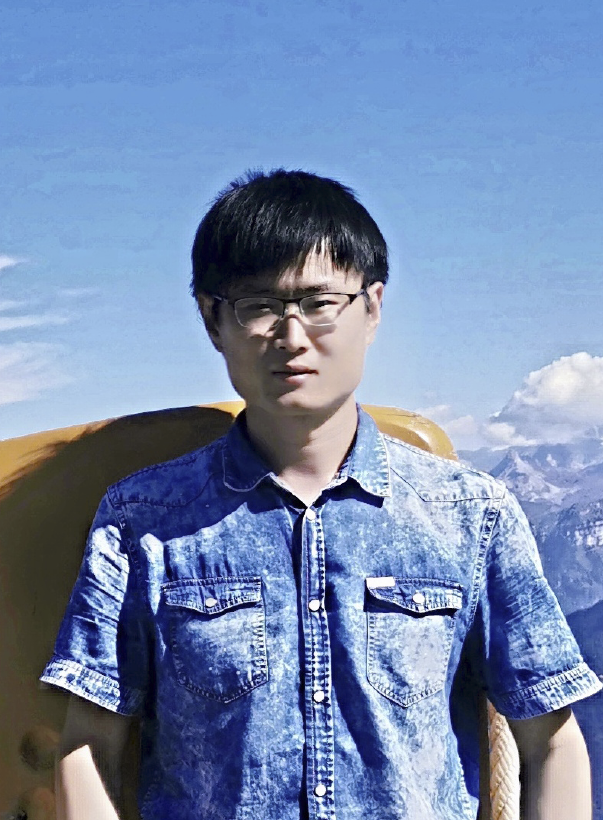}}]{Kai Zhang}
received his Ph.D. degree from School of Computer Science and Technology, Harbin Institute of Technology, China, in 2019. He was a research assistant from July, 2015 to July, 2017 and from July, 2018 to April, 2019 in Department of Computing of The Hong Kong Polytechnic University. He is currently a postdoctoral researcher at Computer Vision Lab, ETH Zurich, Switzerland, working with Prof. Luc Van Gool and Prof. Radu Timofte. His research interests include machine learning and image processing.
\end{IEEEbiography}

\begin{IEEEbiography}[{\includegraphics[width=1in,height=1.25in,clip,keepaspectratio]{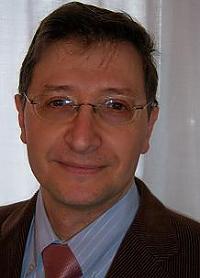}}]{Pietro Li\'o}
received the PhD degree in complex
systems and non linear dynamics from the
School of Informatics, dept of Engineering, University of Firenze, Italy and the PhD degree in
theoretical genetics from the University of Pavia,
Italy. He is currently a professor of computational
biology with the Department of Computer Science and Technology, University of Cambridge
and a member of the Artificial Intelligence Group.
He is also a member of the Cambridge Centre for
AI in medicine, ELLIS (European Laboratory for Learning and Intelligent Systems), Academia Europaea, Asia Pacific Artificial Intelligence Association, . His research interests include graph representation learning,
AI and Medicine, Systems Biology.
\end{IEEEbiography}

\vfill
\end{document}